\documentclass[lettersize,journal]{IEEEtran}
\usepackage{amsmath,amsfonts}
\usepackage{algorithmic}
\usepackage{algorithm}
\usepackage{array}
\usepackage[caption=true,font=footnotesize,labelfont=sf,textfont=sf]{subfig}
\usepackage{textcomp}
\usepackage{stfloats}
\usepackage{url}
\usepackage{verbatim}
\usepackage{graphicx}
\usepackage{cite}

\usepackage[colorlinks,urlcolor=red,linkcolor=red]{hyperref}
\usepackage{booktabs}
\usepackage{mathbbol}
\usepackage{multirow}
\usepackage{makecell}
\usepackage{ulem}

\usepackage[capitalize]{cleveref}
\crefname{section}{Sec.}{Secs.}
\Crefname{section}{Section}{Sections}
\Crefname{table}{Table}{Tables}
\crefname{table}{Tab.}{Tabs.}

\begin{document}

\title{Pushing One Pair of Labels Apart Each Time in Multi-Label Learning: From Single Positive to Full Labels}

\author{Xiang Li$^*$, Xinrui Wang$^*$,
	Songcan Chen
	\IEEEcompsocitemizethanks{\IEEEcompsocthanksitem Xiang Li, Xinrui Wang, and Songcan Chen are with College of Computer Science and Technology/College of Artificial Intelligence, Nanjing University of Aeronautics and Astronautics, Nanjing, 211106, China, and also with MIIT Key Laboratory of Pattern Analysis and Machine Intelligence. Corresponding author is Songcan Chen.\protect\\
	E-mail: \{lx90, wangxinrui, s.chen\}@nuaa.edu.cn.}
	\thanks{Manuscript received February 28, 2023.}
}

\maketitle

\renewcommand{\thefootnote}{\fnsymbol{footnote}}
\footnotetext[1]{Equal contribution.} 


\markboth{Journal of \LaTeX\ Class Files,~Vol.~14, No.~8, August~2021}%
{Shell \MakeLowercase{\textit{et al.}}: A Sample Article Using IEEEtran.cls for IEEE Journals}


\begin{abstract}
	In Multi-Label Learning (MLL), it is extremely challenging to accurately annotate every appearing object due to expensive costs and limited knowledge. When facing such a challenge, a more practical and cheaper alternative should be Single Positive Multi-Label Learning (SPMLL), where only one positive label needs to be provided per sample. Existing SPMLL methods usually assume unknown labels as negatives, which inevitably introduces false negatives as noisy labels. More seriously, Binary Cross Entropy (BCE) loss is often used for training, which is notoriously not robust to noisy labels. To mitigate this issue, we customize an objective function for SPMLL by pushing only one pair of labels apart each time to prevent the domination of negative labels, which is the main culprit of fitting noisy labels in SPMLL. To further combat such noisy labels, we explore the high-rankness of label matrix, which can also push apart different labels. By directly extending from SPMLL to MLL with full labels, a unified loss applicable to both settings is derived. Experiments on real datasets demonstrate that the proposed loss not only performs more robustly to noisy labels for SPMLL but also works well for full labels. Besides, we empirically discover that high-rankness can mitigate the dramatic performance drop in SPMLL. Most surprisingly, even without any regularization or fine-tuned label correction, only adopting our loss defeats state-of-the-art SPMLL methods on CUB, a dataset that severely lacks labels. 
\end{abstract}


\section{Introduction}
\label{sec:intro}
\IEEEPARstart{A}{s} a general extension of multi-class learning, multi-label learning (MLL) \cite {zhang2014survey, gibaja2015tutorial, liu2021emerging} often contains multiple labels in a single training sample, and the goal is to assign every label that associated with the corresponding sample. Since the setting of MLL is much closer to reality where multi-objects often co-occur in a natural scene, it has received considerable attention in the past two decades and has developed wide applications as diverse as image classification \cite {wang2016cnn,zhu2017learning,lanchantin2021general}, video analysis \cite {ray2018scenes,zhang2021multi,ji2021multi}, natural language processing \cite {banerjee2019hierarchical,chen2020hyperbolic,zhang2021fast}, just to name a few. 

As we all know, exhaustively and accurately annotating every object that appears in a sample, \textit{i.e.}, making full labels, is extremely expensive, and sometimes even impossible, since there may exist unknown objects due to limited knowledge. To circumvent the obstacles in making full labels, a good and more practical alternative should be Single Positive Multi-Label Learning (SPMLL) \cite {cole2021multi}, \textit{i.e.}, annotating only one object that appears in a sample and other labels remain unknown or unannotated. 

The advantages of SPMLL lie in the following aspects. First, annotating single positive label is much simpler and less expensive since there is no need to exhaustively examine every corner of the sample. Second, the risk of introducing false positive label is greatly reduced because we tend to annotate the object that we are more familiar with and more confident in. Recently, in \cite {xu2022one}, the authors have proven that single positive label is sufficient for MLL, which provides a theoretical support for this application.

Although SPMLL enjoys the above advantages, naturally, the least annotation of each sample tends to make traditional MLL methods underperform. And those relying on the co-occurrence to learn label correlations \cite {weng2018multi, nguyen2019multi, che2020novel} will even fail due to that the co-occurrence in a single sample is no longer available in the setting of SPMLL. Therefore, in such an emerging field, algorithms specialized for SPMLL should be carefully designed.

\begin{figure*}[htbp]
	\centering
	\includegraphics[width=0.97\linewidth]{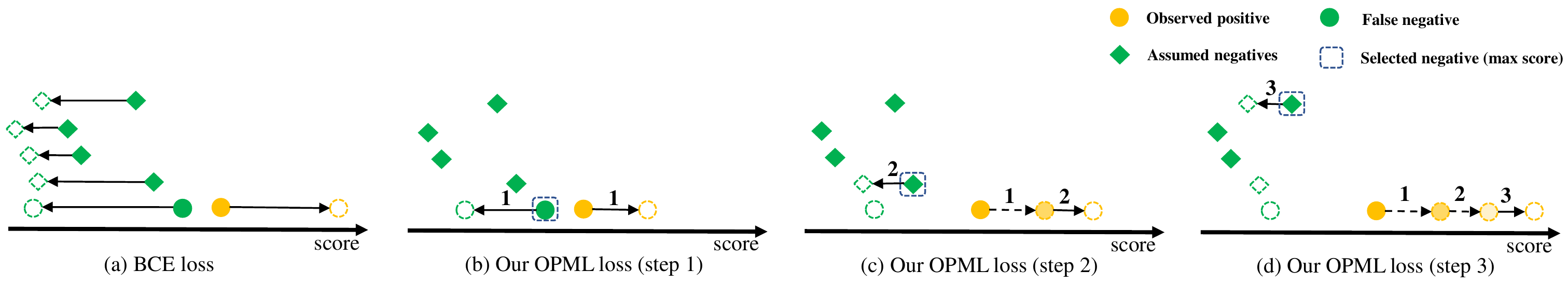} 
	\caption{The optimization procedure for BCE and our OPML loss in SPMLL. The solid and the dotted arrows represent the optimization at current and previous steps, respectively. Rhombuses and circles with solid and dotted lines are the original positions in the initial state and the intermediate positions at current step, respectively. Figure 1a shows that, BCE loss optimizes all pairs of labels at once, thus it can be easily dominated by negative labels in SPMLL. Figure 1b-1d describe three procedures of pushing one pair of labels apart each time. Note that, the selected negative label is the one with the maximum score, \textit{i.e.}, the nearest one to the single positive label at current step.}
	\label{fig1}
\end{figure*}

To deal with the challenge of SPMLL, the pioneering work \cite {cole2021multi} assumes all the unknown labels to be negatives, and the subsequent works \cite{kim2022large, xu2022one, xie2022labelaware} also adopt the so-called ``Assume Negative" (AN) assumption. Obviously, such an assumption inevitably introduces false negatives, which involves noisy labels during training. More seriously, as the default choice of loss function in these works, Binary Cross Entropy (BCE) loss is notoriously not robust to noisy labels \cite {ma2020normalized, wang2021learning, zhou2021learning}. The reason hidden behind is that BCE loss utilizes the same gradient regime for positive and negative labels, which means that BCE loss treats each label equally \cite {zhao2021evaluating, zhou2022acknowledging}. Therefore, BCE loss will fit not only the clean labels but also the noisy labels. Note that in SPMLL, only one positive label is observed for each sample, and the AN assumption makes the number of negative labels far more than that of positive label. Consequently, the negative labels will dominate the training, conversely, the single positive label will be hindered during training, which causes a performance gap between the SPMLL and MLL with full labels.

From the above analyses, it can be seen that there are two key factors for robust SPMLL, one is making good use of the precious single positive label, and the other is preventing the domination of negative labels. To close such gap, in this work, we customize an objective function for SPMLL by pushing only one pair of labels apart each time from the motivation of considering the two key factors. Specifically, we select the observed single positive label and the unobserved label with the maximum score as a pair of labels to be optimized each time, informally, we maximize the difference of the scores between the selected labels to distinguish them. A formal formulation will be detailed in \Cref {sec:approach}. By such a design, only one positive label and one unobserved label are optimized each time, thus the domination of negative labels is prevented and meanwhile the importance of the positive label is emphasized.

For convenience of comparison, a visualized description of the optimization procedure for BCE and our loss is depicted in \Cref {fig1}. In a nutshell, our loss optimizes only one pair of labels each time whereas the BCE loss optimizes all the pairs. From \Cref {fig1}, it can be also found that our loss slows down the process of turning the unknown true positive to false negative. Note that, the strategy of pushing one pair of labels apart can be directly extended to MLL with full labels by selecting the negative label with the maximum score and the positive label with the minimum score as a pair of labels to be optimized. By such an extension, we derive a unified loss for both SPMLL and MLL with full labels.

In SPMLL, as the precise supervised information is seriously insufficient, the unknown true positive label will gradually turn to false negative during training. To further combat the noisy labels, we cooperate the high-rank property of multi-label matrix \cite {li2022concise} with our newly proposed loss. The motivation is that the high-rank property encourages to push apart different labels, which will further slow down the the process of turning the unknown true positive to false negative. Finally, experiments in both SPMLL and MLL with full labels are conducted to verify the effectiveness of our method.

In summary, our contributions are threefold:

(1) We derive a new loss by optimizing One Pair of labels each time for Multi-Label learning abbreviated as OPML, which can be seamlessly used in both the single positive and full labels settings. Experiments show that the OPML loss not only performs more robustly to noisy labels in SPMLL but also works well in MLL with full labels.

(2) We empirically discover that in SPMLL, the high-rank property can alleviate the negative impact of noisy labels during the learning process, specifically, with such regularization, performance drop is not that dramatic, which may shed new light on general noisy label learning.

(3) Compared with state-of-the-art methods \cite {kim2022large,zhou2022acknowledging} that adopt extra regularization or fine-tuned label correction, only adopting OPML loss defeats them on CUB, a dataset that severely lacks labels, even without any of these techniques.

\section{Related Work}
\label{sec:related}

In this section, we review the related works from the following two aspects. The first is the commonly used loss functions in MLL and the mechanisms behind them, the second is recent advances focusing on SPMLL.

\noindent {\textbf{Loss functions in MLL.}} In both MLL and its closely related studies, BCE loss is often used as a default choice combined with various tricks and regularizations \cite {bartlett2008classification, liu2018multi, chen2019multi, gao2020multi}. However, it has been shown that BCE loss is not robust to noisy labels \cite {ma2020normalized, wang2021learning, zhou2021learning}, which are pervasive in MLL. Recently, \cite{ridnik2021asymmetric} has revealed that by alleviating the imbalance problem in MLL, some variants of BCE loss like focal loss \cite{he2017focal} and asymmetric loss \cite{ridnik2021asymmetric} can significantly boost the performance. \cite{he2017focal} is originally proposed to deal with the serious imbalance between the targets and numerous backgrounds in object detection. Since focal loss can well handle the imbalance problem, which is inherent in MLL, it has become a widely used loss in MLL \cite{wu2020distribution, ridnik2021asymmetric, akyurek2020multi, dong2020focal, zheng2021multi} and has achieved great empirical success. \cite{ridnik2021asymmetric} even points out that, by simply reducing the contribution of negative samples to the loss when their probability is low, the BCE loss with the carefully tuned reweighting parameter can reach state-of-the-art results.  Besides, \cite{su2022zlpr} has extended the popular cross-entropy loss to MLL by exploring the proper surrogate function for softmax in MLL, which is a special case fallen into our loss. It is worth mentioning that our proposed OPML loss can be seamlessly applied in both traditional methods and popular deep neural networks.

\noindent {\textbf{Recent advances in SPMLL.}} SPMLL is first proposed by \cite{cole2021multi} considering its significantly reduced annotation costs. Due to the lack of precise supervision, it takes the AN assumption on the unobserved labels combined with various re-weighting strategies by incorporating the expected number of true positive labels, which is not available in reality. Following \cite{cole2021multi}, authors in \cite{kim2022large} also take the AN assumption and cast the SPMLL task into noisy multi-label classification. Later, instead of making the AN assumption, \cite {zhou2022acknowledging} treats all unannotated labels as unknown by maximizing the entropy, and then adopts a heuristic asymmetric pseudo-labeling method. Recently, in \cite {xu2022one}, the authors have proven that single positive label is sufficient for MLL by deriving a risk estimator approximately converging to the optimal risk minimizer of fully supervised learning, which provides a solid theoretical support. Inspired by the empirical success of consistency regularization in multi-class classification, \cite{xie2022labelaware} extends this popular regularization to SPMLL with the help of their proposed label-aware attention module.  Plus, \cite{cho2022mining} studies the setting of SPMLL from the perspective of generating multi-label data from a single positive label. Despite great efforts have been made, the performance gap between the SPMLL and MLL with full labels still exists.

\section{Proposed approach}
\label{sec:approach}
\subsection{Problem statement}
In this subsection, we first describe the setting of MLL with full labels, and then detail the setting of SPMLL.

\noindent {\textbf{MLL with full labels.}} Given a training dataset $\mathcal{D} = \left \{{X_i,Y_i} \right \}_{i=1}^{N}$, where $N$ is the number of training samples. ${Y}_i\in \left \{ 0,1 \right \}^{L}$, $Y_{il}=1$ if the $l$-th label is relevant to $X_i$, which is also called positive label, otherwise $Y_{il}=0$, also knowing as negative label ($l\in \left \{ 1,2,\cdots ,L \right \} $), where $L$ is the number of labels. All the labels of each ${Y}_i$ ($i\in \left \{ 1,2,\cdots ,N \right \} $) in the training set are observed under the setting of MLL with full labels. The goal is to train a model $f : \mathcal{X} \longmapsto \left [ 0,1 \right ]^{L} $ that outputs the labels of unseen samples in the feature space $\mathcal{X}$. 

\noindent {\textbf{SPMLL.}} For SPMLL, there is only one label observed for each ${Y}_i$ ($i\in \left \{ 1, 2, \cdots, N \right \} $) in the training set. Formally, $Y_{il}$ satisfies the following two conditions. (1) $Y_{il}\in \left \{ 1, \emptyset \right \}$ for all $i\in \left \{ 1,2,\cdots ,N \right \} $ and $l\in \left \{ 1,2,\cdots ,L \right \} $, where $Y_{il} = \emptyset$ denotes that the $l$-th label in the $i$-th sample is unobserved. (2) ${\textstyle \sum_{l=1}^{L} \mathbb{1}_{\left [{Y}_{il}=1\right ]} =1 }$ for all $i\in \left \{ 1,2,\cdots ,N \right \} $, where $\mathbb{1}_{\left [\bullet \right ]}$ is the indicator function, which equals to $1$ when the proposition in the square brackets holds, and $0$ otherwise \cite {cole2021multi}. Obviously, in the SPMLL, the least annotation of each sample makes the supervised information severely insufficient, which causes performance gap between the SPMLL and MLL with full labels.   

\subsection{The proposed OPML loss}
To close such performance gap, existing SPMLL methods \cite {cole2021multi, kim2022large, xu2022one, xie2022labelaware} have made great efforts on it. Most of them assume the unknown labels as negative labels, which inevitably introduces false negatives as noisy labels. More seriously, as the commonly used loss functions in these works, BCE loss is notoriously not robust to noisy labels \cite {ma2020normalized, wang2021learning, zhou2021learning}. To prevent the domination of negative labels, which is the main culprit of fitting noisy labels in SPMLL, we tailor an objective function to push only one pair of labels apart each time. Concretely, we select the observed single positive label and the unobserved label with the maximum score as the pair of labels, and then maximize the difference of the scores between the selected pair of labels to distinguish them. Formally, it can be formulated as follows: 
\begin{equation}
\max_{\theta} \left ( s_{p}  - \max_{n\in \Omega_{n}} \ s_{n}\right ),
\label{eq:1}
\end{equation}
where $\theta$ is the parameter of deep neural networks, $s_{p}$ and $s_{n}$ are the scores of single positive label and assumed negative labels, respectively, and $\Omega_{n}$ is the index set of assumed negative labels. Note that, the maximum function is non-differentiable, thus we employ the smooth approximation of maximum function $logsumexp$ \cite {blanchard2021accurately}, which is defined as $logsumexp(x_1,x_2, \cdots, x_k)=log {\textstyle \sum_{i=1}^{k}}  e^{(x_i)}$ for any $x_i \in \left ( -\infty, \infty \right )$ and $i={1,2, \cdots, k}$, where $log(\bullet)$ is the natural logarithmic function, and $e^{(\bullet)}$ is the exponential function. 
\begin{figure*}[htbp]
	\centering
	\includegraphics[width=0.9\linewidth]{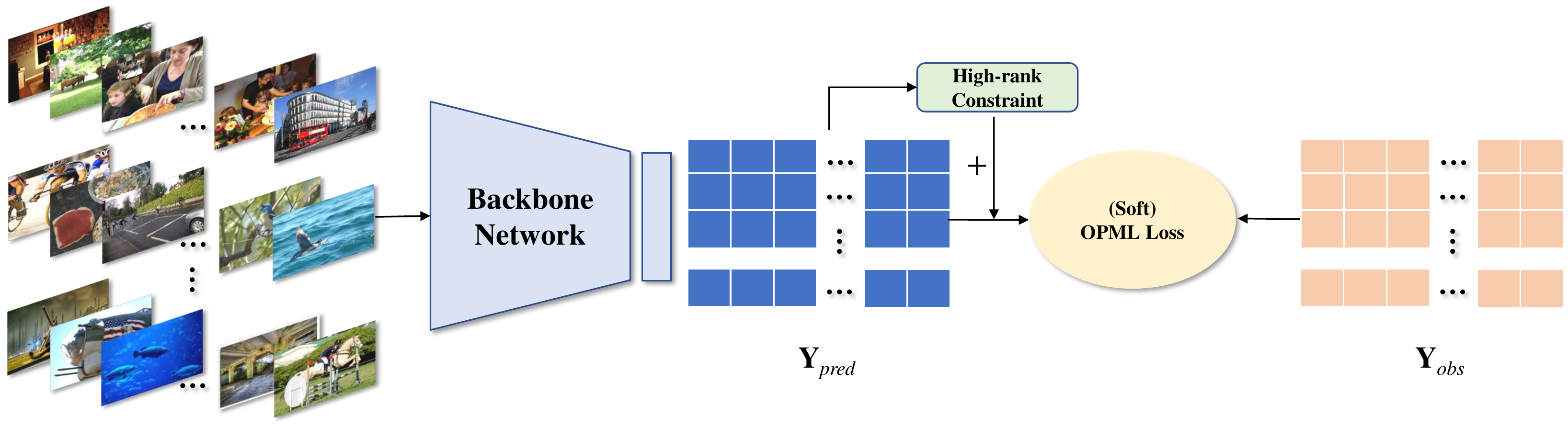} 
	\caption{The framework of cooperating the high-rank regularization with our OPML loss, where $\mathbf{Y}_{pred}$ and $\mathbf{Y}_{obs}$ denote the predicted and observed label matrices, respectively.}
	\label{fig2}
\end{figure*}

By the smooth approximation of $logsumexp$ and some simple mathematical computations, we have the following objective function:
\begin{equation}
\min_{\theta} \left ( -s_{p}  + log \sum_{n \in \Omega_n}  e^{s_n} \right ).
\label{eq:3}
\end{equation}

Let $\mathcal{L} =  -s_{p}  + log \textstyle \sum_{n \in \Omega_n}  e^{s_n} = log \ e^{(-s_p)}  + log \textstyle \sum_{n \in \Omega_n}  e^{s_n}$ be the loss function for SPMLL. Note that, this loss is unbounded and the optimal tends to negative infinity, which is unstable and difficult to be optimized in deep neural networks. To make this loss function bounded and stable, we add two positive constants $\alpha$ and $\beta$ into $\mathcal{L}$ and achieve the final loss function customized for SPMLL, denoted as $\mathcal{L}_{SPOPML}$,
\begin{equation}
\mathcal{L}_{SPOPML} = log(\alpha + e^{(-s_p)})  + log( \beta + \sum_{n \in \Omega_n}  e^{s_n}).
\label{eq:44}
\end{equation}

Note that, the objective function in \cref {eq:1} can be easily extended to MLL with full labels by selecting the negative label with the maximum score and the positive label with the minimum score as one pair of labels to be optimized. Similarly, the formal formulation can be written as: 
\begin{equation}
\max_{\theta} \left ( \min_{p\in \Omega_{p}} \ s_{p}  - \max_{n\in \Omega_{n}} \ s_{n}\right ),
\label{eq:4}
\end{equation}
where $\Omega_{p}$ and $\Omega_{n}$ are the index sets of positive and negative labels, respectively. Note that $\min (a,b) = -\max (-a,-b)$, then we can equivalently transform \cref {eq:4} into the following formulation:
\begin{equation}
\max_{\theta} \left ( -\max_{p\in \Omega_{p}} (-s_{p})  - \max_{n\in \Omega_{n}} \ s_{n}\right ).
\label{eq:5}
\end{equation} 
For conciseness, similar to the transformation of \cref {eq:1} to \cref {eq:44}, we achieve the final loss function for MLL with full labels. 
\begin{equation}
\mathcal{L}_{OPML} = log(\alpha + \sum_{p \in \Omega_p}  e^{(-s_p)})  + log( \beta + \sum_{n \in \Omega_n}  e^{s_n}).
\label{eq:7}
\end{equation}

It is worth emphasizing that \cref {eq:7} degenerates to \cref {eq:44} when only single positive label is observed. Hence, by \cref {eq:7}, a unified loss for both SPMLL and MLL with full labels is derived.

Since $\alpha$ and $\beta$ belong to $(0, \infty)$, which are too wide to select appropriate parameters, here we provide a simple yet flexible mechanism to choose them. Specifically, let $\alpha =  \widetilde{\alpha} / (1 - \widetilde{\alpha})$ and $\beta =  \widetilde{\beta} / (1 - \widetilde{\beta})$, where $\widetilde{\alpha}$ and $\widetilde{\beta}$ belong to $(0, 1)$. By this transformation, we can choose the parameters from $(0, 1)$ while still keeping $\alpha$ and $\beta$ belong to $(0, \infty)$.

\subsection{Gradient analysis}

In this section, to better understand why BCE loss fits the noisy labels and why our OPML loss performs more robustly than BCE loss in SPMLL, we make detailed gradient analyses for both of them. Given a data point ($X_{i}$, $Y_{i}$) for any $i \in \{ 1,2,\cdots,N \}$, the BCE loss is 
\begin{equation}     \label{eqS1}
\begin{aligned}
\mathcal{L}_{\mathrm{BCE}} = & -\frac{1}{L} \sum_{l=1}^{L} {\left[\mathbb{1}_{\left[Y_{i l}=1\right]} \log \left(f_{i l}\right)\right.} \\
& \left.+\mathbb{1}_{\left[Y_{i l}=0\right]} \log \left(1-f_{i l}\right)\right]
\end{aligned}
\end{equation}
where $N$ and $L$ are the numbers of training samples and labels, respectively. $Y_{i l}$ is the $l$-th label of the $i$-th sample, $f_{i l}$ is the corresponding predicted score. For the brevity of notation, let $f = f_{i l}$, since the sigmoid function $g = 1/(1 + e^{(-f)})$ is used as the activation function in the BCE loss, then the gradient of BCE loss with respect to the score $f$ can be written as:
\begin{equation}     \label{eqS2}
\left\{\begin{array}{ll}
\frac{\partial \mathcal{L}_{BCE}}{\partial f}=\frac{\partial \mathcal{L}_{BCE}}{\partial g} \frac{\partial g}{\partial f}=\frac{-e^{-f}}{1+e^{-f}}, & Y_{i l}=1 \\
\frac{\partial \mathcal{L}_{BCE}}{\partial f}=\frac{\partial \mathcal{L}_{BCE}}{\partial g} \frac{\partial g}{\partial f}=\frac{e^{f}}{1+e^{f}}, & Y_{i l}=0
\end{array}\right.
\end{equation}

Note that, this gradient is a centrosymmetric function about the origin, thus BCE loss utilizes the same gradient regime for positive and negative labels, which means that BCE loss treats each label equally \cite {zhou2022acknowledging}. Consequently, BCE loss will fit not only the clean labels but also the noisy labels during training.

By taking the derivative of \cref {eq:7}, the gradients of our OPML loss with respect to the scores $s_p$ and $s_n$ in the SPMLL setting can be calculated as:
\begin{equation}     \label{eqS4}
\left\{\begin{array}{ll}
\frac{\partial \mathcal{L}_{OPML}}{\partial s_p} = \frac{-e^{-s_p}}{\alpha+\sum_{p \in \Omega_p}  e^{(-s_p)}} = \frac{-e^{-s_p}}{\alpha+ e^{(-s_p)}}, & Y_{i l}=1 \\
\frac{\partial \mathcal{L}_{OPML}}{\partial s_n} = \frac{e^{s_n}}{\beta+\sum_{n \in \Omega_n}  e^{s_n}}, & Y_{i l}=0
\end{array}\right.
\end{equation}

For ease of comparison, let $\alpha = 1$ and $\beta = 1$, then we can see that, the gradient of OPML loss with respect to the score of the single positive label is the same as that of BCE loss while the gradient of OPML loss with respect to the score of the negative labels is less than that of BCE loss due to the fact that ${e^{f}}/({1+e^{f}}) > {e^{f}}/({1+\sum_{} e^{f}})$. In conclusion, compared with BCE loss, the domination of negative labels is alleviated and meanwhile the importance of the positive label is maintained in SPMLL by adopting the OPML loss for training.

\subsection{High-rank regularization}
As mentioned in \Cref {sec:intro}, in SPMLL, the unknown true positive label will gradually turn to false negative during training. Motivated by the intuition that different labels fall into different subspaces, the label matrix is often prone to be high-rank \cite {li2022concise}. To further reduce the risk brought by noisy labels, we explore such high-rank property to push apart different labels, which aims at further slowing down the the process of turning the unknown true positive to false negative. Specifically, we add a high-rank constraint on the prediction label matrix, which can be formulated as follows:
\begin{equation}
\mathcal{R}_{HR}=-\lambda \log \operatorname{det}\left(\mathbf{Y}_{pred}^{T} \mathbf{Y}_{pred}\right)=- \lambda \sum_{i} \log (\sigma_{i}^{2}),
\label{eq:8}
\end{equation}  
where $\lambda$ is a trade-off hyper-parameter, $\operatorname{det}$ is the determinant function, $\mathbf{Y}_{pred}$ is the predicted label matrix, and $\sigma_{i}$ is its corresponding singular value. Here, we adopt the minus $\operatorname{logdet}$ function \cite {fazel2003log} as the high-rank regularization rather than the minus trace norm used in \cite {li2022concise} for the reason that the $\operatorname{logdet}$ function is differentiable, which is easier to optimize in deep neural networks. The framework of cooperating this high-rank regularization with our newly proposed OPML loss is depicted in \Cref {fig2}.

\subsection{Soft variant and label correction}
As our OPML loss is customized for SPMLL, a setting with unavoidable noisy labels under the AN assumption, then it is natural to ask whether our OPML loss can be combined with some commonly used techniques in noisy label learning, such as label smoothing \cite {lukasik2020does,zhang2021delving,lukov2022teaching} and label correction \cite {liu2020early, kim2021fine, zheng2021meta}. To answer this question, in this subsection, we propose a soft variant of OPML loss by label smoothing and further combine it with a label correction mechanism to verify that our loss can be well cooperated with these techniques.

\noindent {\textbf{Soft OPML loss.}} In SPMLL, most methods assume the unknown labels as negatives, which incorrectly annotates the unobserved positives as false negatives. To combat such noisy labels, we utilize the label smoothing to soften the hard discrete labels $\{0,1\}$ to continuous labels $[0,1]$. Formally, the soft variant of OPML loss can be rewritten as:
\begin{equation}
\begin{aligned}
\mathcal{L}_{SOPML} 
& = \log(\alpha + e^{(-s_p)}) + \log(\alpha +\sum_{l \in \mathcal{U}} \gamma_{l} e^{(-s_l)}) \\
& + \log(\beta+\sum_{l \in \mathcal{U}}\left(1-\gamma_{l}\right) e^{s_l}),
\end{aligned}
\label{eq:9}
\end{equation} 
where $\mathcal{U}$ is the index set of unobserved labels, $\gamma_{l}$ is a smoothing parameter of the $l$-th label. It is worth noting that instead of manual selection, $\gamma_{l}$ is a dynamic adaptive parameter computed by the metric of Average Precision (AP) \cite {wu2017unified} on the training set. Concretely, $\gamma_{l} = pred_{l} \times AP_{l}^{\epsilon}$, where $\epsilon$ is a power parameter, $pred_{l}$ and $AP_{l}$ are the prediction score and AP score of the $l$-th label, respectively. The intuition of calculating the adaptive parameter in such a way is that the larger the AP, the more reliable its corresponding prediction score is.

\noindent {\textbf{Label correction.}} In noisy label learning, label correction \cite {liu2020early, kim2021fine, zheng2021meta} is a common and important data cleansing technique. To further validate that our loss can also be well combined with the label correction technique, we likewise propose an AP-based label correction mechanism. Our motivation is intuitive that the larger the AP, the less number the corresponding label is modified. For each class of label, the number of labels to be corrected can be calculated as $Cornum_{l} = Tr_{num} \times Cor_{ratio} \times (1 - AP_{l})$, where $Cornum_{l}$ is the number of labels to be corrected for the $l$-th label, $Tr_{num}$ is the number of training samples, $Cor_{ratio}$ is a parameter of label correction ratio, $AP_{l}$ is the AP score of the $l$-th label. Note that we use the AP scores of the observed labels in our label smoothing and correction mechanisms.The process of the label correction is summarized in the following algorithm.

\begin{algorithm} [h] \label{Alg1}  
	\caption{The process of label correction}
	
	\textbf{Input}: Number of training samples $Tr_{num}$, label correction ratio $Cor_{ratio}$, training average precision score vector $AP$.\\
	\textbf{Output}: The modified label matrix $Y_{m}$.
	
	\begin{algorithmic}[1] 
		\FOR {$l=1$ to $L$}
		\STATE Computing the number to be corrected, $Cornum_{l} = Tr_{num} \times Cor_{ratio} \times (1 - AP_{l})$.
		\STATE Sorting the score of unobserved label $Score_{u}$ in descending order.
		\STATE Changing the negative labels of the first $Cornum_{l}$-th samples to positives.
		\ENDFOR
		\STATE \textbf{return} the modified label matrix $Y_{m}$.
	\end{algorithmic}
\end{algorithm}

\noindent {The $Cor_{ratio}$ in step 2 is computed by $Cor_{ratio} = Obsnum_{l} / Tr_{num} \times Label_{num}$, where $Obsnum_{l}$ is the number of observed labels in the $l$-th label, and $Label_{num}$ is a parameter that will be specified in the \Cref{exp}.}
\section{Experiments}
\label{exp}
In this section, we conduct extensive experiments on multi-label image classification in both single positive label and full labels settings to verify the effectiveness of our proposed method.
\subsection{Experiments settings}
\noindent {\textbf {Datasets.}} For fairness of comparisons, we follow the setting in \cite {cole2021multi, kim2022large, zhou2022acknowledging}. Specifically, four standard benchmark datasets, PASCAL VOC 2012 (VOC) \cite {voc2012}, MS-COCO 2014 (COCO) \cite {lin2014microsoft}, NUS-WIDE (NUS) \cite {chua2009nus}, and CUB-200-2011 (CUB) \cite {wah2011caltech} are used for evaluation. Given four full labels datasets, first, for each dataset, $20\%$ of the training set is withheld for validation. Second, to create single positive training data, one positive label is randomly selected to keep for each training sample, and the remaining annotations are discarding, which is performed only once for each dataset and the single positive training set is fixed. The statistics of four datasets are shown in \Cref {tab1}. Note that, we use totally the same number of seeds as in \cite {cole2021multi, kim2022large, zhou2022acknowledging} for random sampling and spitting to create the same dataset for fair comparisons.

\begin{table}[!ht]
	\centering
	\small
	\setlength\tabcolsep{7pt}
	{
		\begin{tabular}{ccccc}
			\toprule[1.5pt]
			Datasets  & \#Class    & \#Training & \#Validation & \#Test   \\ 
			\midrule[0.8pt]
			VOC      & 20     & 4574    & 1143    & 5823         \\
			COCO        & 80     & 65665   & 16416   & 40137        \\
			NUS     & 81     & 120000  & 30000   & 60260        \\ 
			CUB         & 312    & 4795    & 1199    & 5794         \\       
			\bottomrule[1.5pt]
		\end{tabular}
	}
	\caption{Statistics of the Datasets.}
	\label{tab1}
\end{table}

\begin{table*}[!ht]
	\centering
	\small
	\setlength\tabcolsep{12pt}
	{
		\begin{tabular}{cccccc}
			\toprule[1.5pt]
			Observed labels           & Methods & VOC & COCO & NUS & CUB \\ \midrule[0.8pt]
			All P. \& All N.         & BCE    & 89.42(0.27) & 76.78(0.13) & 52.08(0.20) & 30.90(0.64) \\
			1 P. \& All N.           & BCE    & 87.60(0.31) & 71.39(0.19) & 46.45(0.27) & 20.65(1.11) \\ \midrule[0.8pt]
			\multirow{13}{*}{1 P. \& 0 N. }       
			& BCE & 85.89(0.38) & 64.92(0.19) & 42.27(0.56) & 18.31(0.47) \\
			& BCE+DW & 86.98(0.36) & 67.59(0.11) & 45.71(0.23) & 19.15(0.56) \\
			& BCE+L1R & 85.97(0.31) & 64.44(0.20) & 42.15(0.46) & 17.59(1.82) \\ 
			& BCE+L2R & 85.96(0.36) & 64.41(0.24) & 42.72(0.12) & 17.71(1.79) \\
			& BCE+LS & 87.90(0.21) & 67.15(0.13) & 43.77(0.29) & 16.26(0.45) \\ 
			& BCE+N-LS & 88.12(0.32) & 67.15(0.10) & 43.86(0.54) & 16.82(0.42) \\ 
			& ROLE (CVPR21) & 87.77(0.22) & 67.04(0.19) & 41.63(0.35) & 13.66(0.24) \\ 
			& ROLE+LI (CVPR21) & 88.26(0.21) & 69.12(0.13) & 45.98(0.26) & 14.86(0.72) \\
			& LL-R (CVPR22) & 88.79(0.03) & 70.36(0.21) & 48.10(0.12) & 20.55(0.18) \\ 
			& LL-Ct (CVPR22) & 88.80(0.11) & 70.27(0.08) & \underline{48.18(0.17)} & 20.53(0.18) \\ 
			& LL-Cp (CVPR22) & 88.37(0.23) & 70.37(0.13) & 47.92(0.02) & 20.55(0.22) \\  
			& BCE+EntMax (ECCV22)& 89.09(0.17) & 70.70(0.31) & 47.15(0.11) & 20.85(0.42) \\
			& BCE+EntMax+APL (ECCV22) & \underline{89.19(0.31)} & \underline{70.87(0.23)} & 47.59(0.22) & 21.84(0.34) \\ \midrule[0.8pt]
			\multirow{2}{*}{1 P. \& 0 N. }          
			& OPML (Ours) & 87.77(0.04) & 68.93(0.06) & 47.93(0.05) & \underline{22.30(0.08)} \\
			& OPML-SP (Ours) & \textbf{89.20(0.03)} & \textbf{71.75(0.07)} & \textbf{50.14(0.09)} & \textbf{24.11(0.22)}  \\ \bottomrule[1.5pt]    
		\end{tabular}
	}
	\caption{Compared results of the methods customized for SPMLL and our proposed methods on four SPMLL benchmarks. The mean and standard deviation of mAP are reported. All P. \& All N. denotes that all labels are observed, \textit{i.e.}, the setting of full labels, $1$ P. \& All N. means that one positive and all the negative labels are observed, and $1$ P. \& $0$ N. signifies that only one positive label is observed while others remain unknown, \textit{i.e.}, the setting of single positive. The best result is in bold and the second best result is underlined.}
	\label{tab2}
\end{table*}

\noindent {\textbf {Compared methods.}} For SPMLL, we compare our method with two kinds of methods, one is the methods customized for SPMLL, and the other is the commonly used loss functions in MLL. The latter contains BCE loss, focal loss \cite {he2017focal}, asymmetric loss \cite {ridnik2021asymmetric}, and ZLPR loss \cite {su2022zlpr}, and the former includes BCE + DW (down-weighting negative labels), BCE + L1R/L2R ($l_1/l_2$ regularization), BCE + LS (label smoothing), BCE + N-LS (label smoothing for only negative labels), ROLE (regularized online label estimation) \cite {cole2021multi}, ROLE + LI (ROLE combined with the ``LinearInit'') \cite {cole2021multi}, LL-R (large loss with rejection) \cite {kim2022large}, LL-Ct (large loss with temporary correction) \cite {kim2022large}, LL-Cp (large loss with permanent correction) \cite {kim2022large}, BCE + EntMax (entropy maximization regularization) \cite {zhou2022acknowledging}, and BCE + EntMax + APL (BCE + entropy maximization regularization + asymmetric pseudo-labeling) \cite {zhou2022acknowledging}. 

\noindent {\textbf {Implementation details.}}
Following \cite {cole2021multi, kim2022large, zhou2022acknowledging}, ResNet-50 \cite {he2016deep} pretrained on the ImageNet \cite {russakovsky2015imagenet} is adopted as the backbone network in all the experiments. Although simply using the up-to-date weights of ResNet-50 pretrained on the ImageNet can achieve better performance, we still use the old V1 version as the compared methods for fair comparisons. The grid search is adopted for model selection, and the best hyper-parameters are selected by the best mean average precision (mAP) on the validation set. Each experiment runs three times, and both the mean and standard deviation of mAP are reported. Due to the limited space, the best parameters for each dataset are provided in the supplementary materials.

\subsection {Experimental results of SPMLL}
\label{sec4.2}

In this subsection, we report the results of our methods and the compared methods by conducting experiments on four SPMLL benchmarks, and then make detailed analyses.

In \Cref {tab2}, BCE loss in SPMLL setting can be seen as a baseline, and methods starting from ``ROLE" focus on SPMLL. We also report the results of BCE loss with full labels as an oracle. Our method OPML-SP is short for OPML loss for Single Positive setting, which is the soft OPML loss with high-rank regularization and label correction. From \Cref {tab2}, we have the following findings. 1) Compared with full labels, the mAP of BCE in SPMLL drops dramatically, \textit{e.g.}, $11.86\%$ and $12.59\%$ decrease on COCO and CUB, respectively, which illustrates that BCE loss is not appropriate for SPMLL. 2) Although some of the BCE variants improve the performance, there still exists a large gap between the SPMLL and full labels, especially on CUB, which severely lacks labels in SPMLL. Besides, the second best result is not obtained by the same method, which indicates that the compared methods are not suitable for all the four datasets. Whereas our OPML-SP performs the best on all four datasets and closes the performance gap between SPMLL and full labels, \textit{e.g.}, there are only $0.22\%$ and $1.94\%$ differences on VOC and NUS, respectively. And on the most difficult dataset CUB, OPML-SP achieves a new state-of-the-art result, which demonstrates that our method performs more robustly than the compared methods even when there severely lacks labels. 3) Most surprisingly, even in the SPMLL setting, OPML-SP still performs better than BCE in the one positive and all negative labels observed setting. Besides, only adopting the OPML loss defeats state-of-the-art SPMLL methods with regularizations or fine-tuned label correction on CUB. These two funny results can be attributed to that our strategy of pushing one pair of labels apart each time not only prevents the domination of negative labels but also emphasizes the importance of the observed single positive label.
\begin{table}[h]
	\centering
	\small
	\setlength\tabcolsep{3pt}
	{
		\begin{tabular}{ccccc}
			\toprule[1.5pt]
			Methods & VOC & COCO & NUS & CUB \\ \midrule[0.8pt] 
			BCE & 85.89(0.38) & 64.92(0.19) & 42.27(0.56) & 18.31(0.47) \\ 
			Focal & 87.59(0.58) & \underline{68.79(0.14)} & 47.00(0.14) & 19.80(0.30) \\ 
			ASL & \underline{87.76(0.51)} & 68.78(0.32) & 46.93(0.30) & 18.81(0.48) \\
			ZLPR & 87.63(0.03) & 68.41(0.24) & \underline{47.72(0.14)} & \underline{21.02(0.12)} \\ \midrule[0.8pt]
			OPML & \textbf{87.77(0.04)} & \textbf{68.93(0.06)} & \textbf{47.93(0.05)} & \textbf{22.30(0.08)} \\  \bottomrule[1.5pt]
		\end{tabular}
	}
	\caption{Quantitative results of the commonly used loss functions and our OPML loss on four SPMLL benchmarks. The mean and standard deviation of mAP are reported. The best result is in bold and the second best result is underlined.}
	\label{tab3}
\end{table}

\begin{table*}[t] 
	\centering
	\small
	\setlength\tabcolsep{11pt}
	{
		{
			\begin{tabular}{cccccccc}
				\toprule[1.5pt]
				$\mathcal{L}_{OPML}$  &$\mathcal{R}_{HR}$       & Smoothing  &  Correction   & VOC           & COCO              & NUS            & CUB               \\ \midrule[0.8pt]
				\checkmark  &               &            &               & 87.77(0.04) & 68.93(0.06)  & 47.93(0.05)   & 22.30(0.08)  \\
				\checkmark  &\checkmark     &            &               & 87.81(0.11) & 69.36(0.02)  & 48.11(0.06)   & 22.41(0.10)  \\
				\checkmark  &               & \checkmark &               & 88.62(0.13) & 70.42(0.13)  & 48.98(0.20)   & 22.37(0.06)  \\
				\checkmark  &               & 			 & \checkmark    & 88.36(0.10) & 69.38(0.18)  & 48.97(0.07)   & 23.62(0.18)  \\ 
				\checkmark  &\checkmark     & \checkmark & \checkmark    & \textbf{89.20(0.03)} & \textbf{71.75(0.07)}  & \textbf{50.14(0.09)}   & \textbf{24.11(0.22)}  \\ 
				\bottomrule[1.5pt]
			\end{tabular}
		}
	}
	\caption{The results of ablation study on four SPMLL benchmarks. The $\checkmark$ means with the corresponding component. The mean and standard deviation of mAP are reported. The best results are in bold.}
	\label{tab4}
\end{table*}

In \Cref {tab3}, results of the commonly used loss functions and our OPML loss are displayed. From this table, we can find that the performance of BCE loss is the worst, which also verifies that BCE loss is not appropriate for SPMLL. Although the variants of BCE loss, \textit{i.e.}, focal and asymmetric losses improve the performance compared with BCE loss, they are still not good enough. Moreover, on the dataset CUB, the improvements are marginal, the reason for this result is that in the setting of SPMLL, CUB suffers from a severe lack of labels, making it more susceptible to noisy labels. Besides, as a special case of our OPML loss, ZLPR achieves a inferior performance, which validates that OPML loss is more flexible and has more potential in SPMLL.

\begin{figure} [h]
	\centering
	\subfloat [Histogram on VOC.] {\label{fig33a} \includegraphics[width=0.47\linewidth] {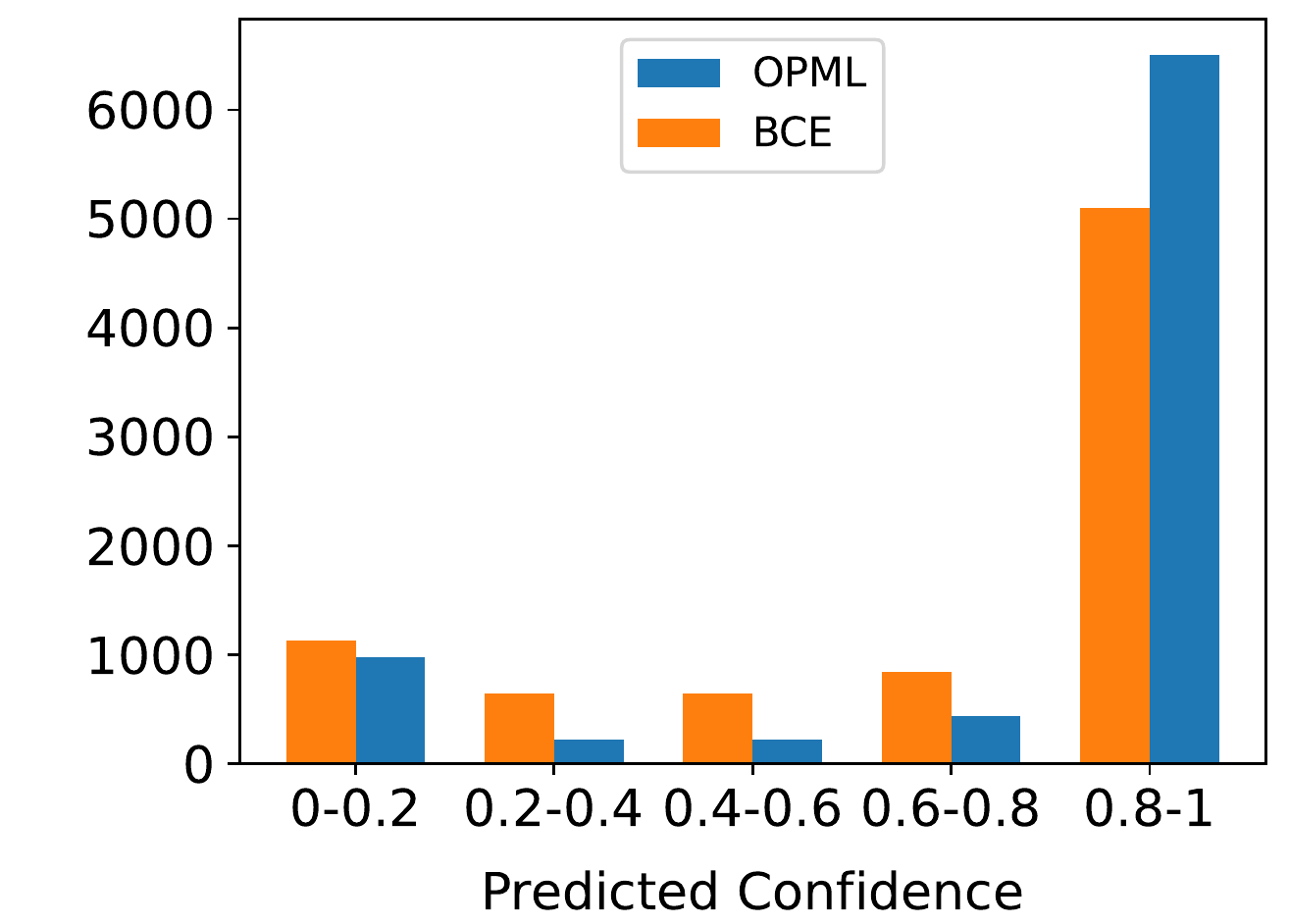}}
	\subfloat [Histogram on CUB.] {\label{fig33b} \includegraphics[width=0.47\linewidth] {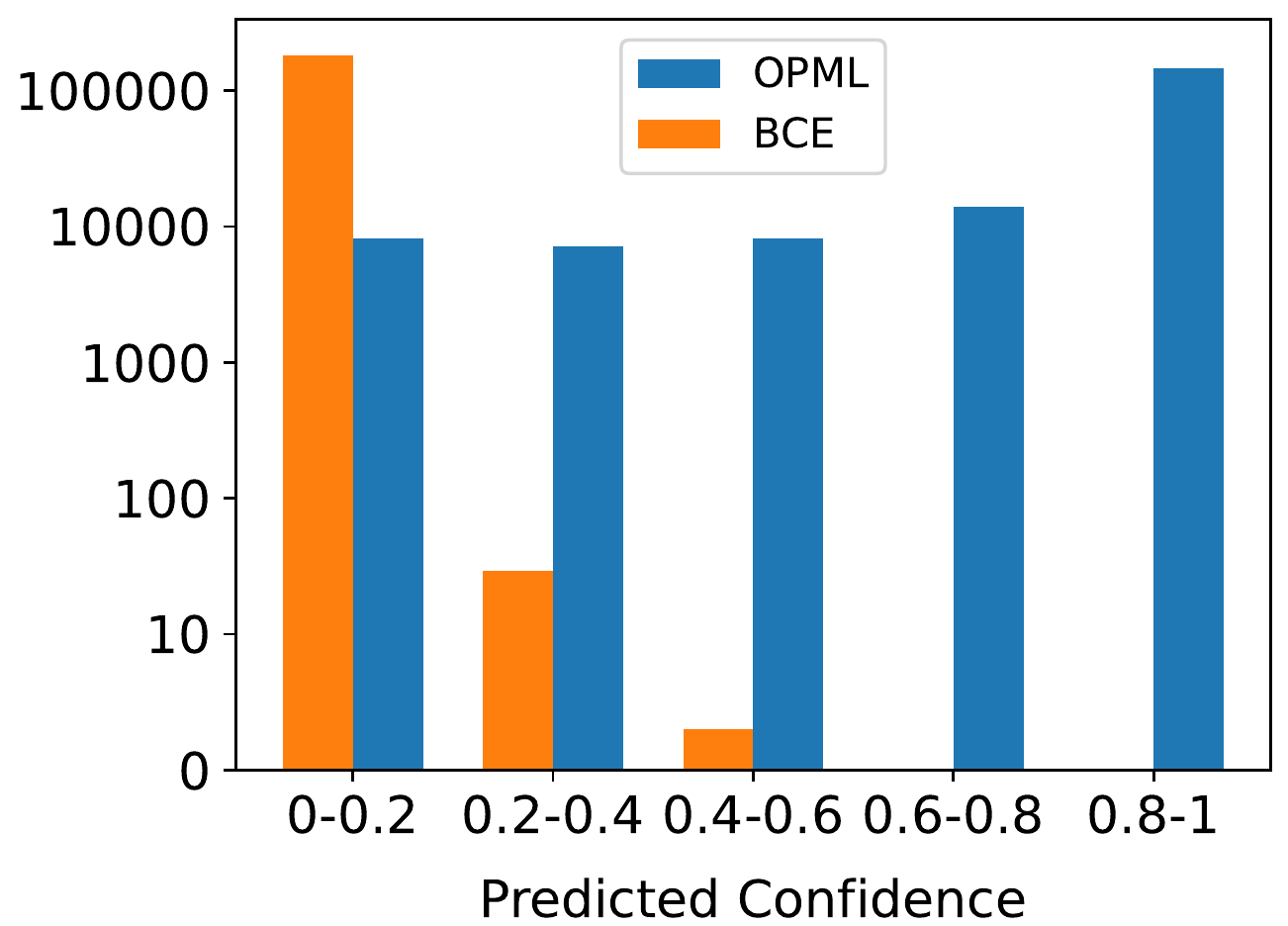}}
	\caption{The histogram of the predicted confidence corresponding to the positive labels in the test label matrix. The ``OPML" with blue and ``BCE" with orange denote the histogram of the predicted confidence trained with OPML and BCE loss, respectively. The width of each bin is $0.2$, note that in \Cref{fig33b}, when trained with BCE loss, the frequencies of the predicted confidence that falling into the interval of $[0.6,0.8)$ and $[0.8,1]$ are zero.} 
	\label{fig33}
\end{figure}

In the following, we conduct statistical experiments to see the distribution of predicted confidence with respect to the positive labels for test set. First, we find the positions where the ground truth in the test label matrix are positive labels, then correspondingly, the predicted confidence scores at the same position in the predicted label matrix are selected for statistical experiments. The \Cref {fig33} shows the statistical histogram of the predicted confidence corresponding to the positive labels in the test label matrix. From \Cref {fig33a}, we can find that in the interval of $[0.8,1]$, the frequency of OPML is larger than BCE, which means when trained with the OPML loss, the predicted confidence scores of positive labels are more inclined to $1$. Besides, in the interval of $[0,0.2]$, the frequency of OPML is smaller than BCE, indicating that OPML can alleviate the domination of negative labels in SPMLL. This phenomenon is more obvious in \Cref {fig33b}, when trained with BCE loss, the frequencies of the predicted confidence that falling into the interval of $[0.6,0.8]$ and $[0.8,1]$ are zero, which means that compared with BCE loss, our OPML loss has a more prominent advantage when the dataset is with more unannotated positive labels. 

\subsection{Ablation study}
\label{sec4.3}
In this subsection, we conduct experiments to verify the effectiveness of the high-rank regularization. Besides, we also validate that our OPML loss can be well cooperated with label smoothing and label correction, which are commonly used techniques in noisy label learning. Unless otherwise specified, the experiments in \Cref {sec4.2,sec4.3,sec4.4,sec4.5} are all performed in the setting of single positive.

\begin{figure} [h]
	\centering	
	\subfloat [mAP of 40 epochs on VOC.] {\label{fig3a} \includegraphics[width=0.473\linewidth]{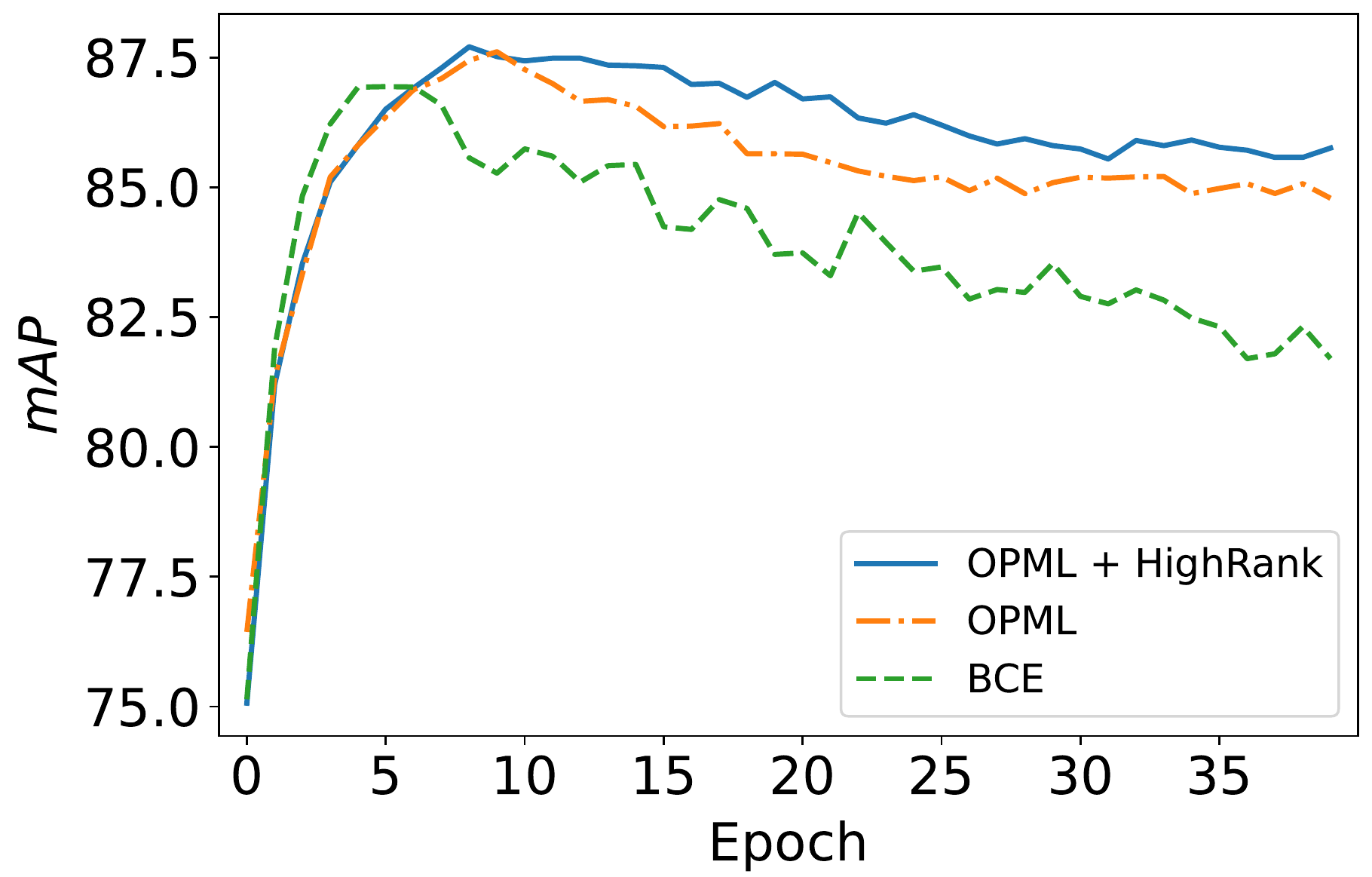}}				 
	\subfloat [mAP of 40 epochs on CUB.] {\label{fig3b} \includegraphics[width=0.473\linewidth]{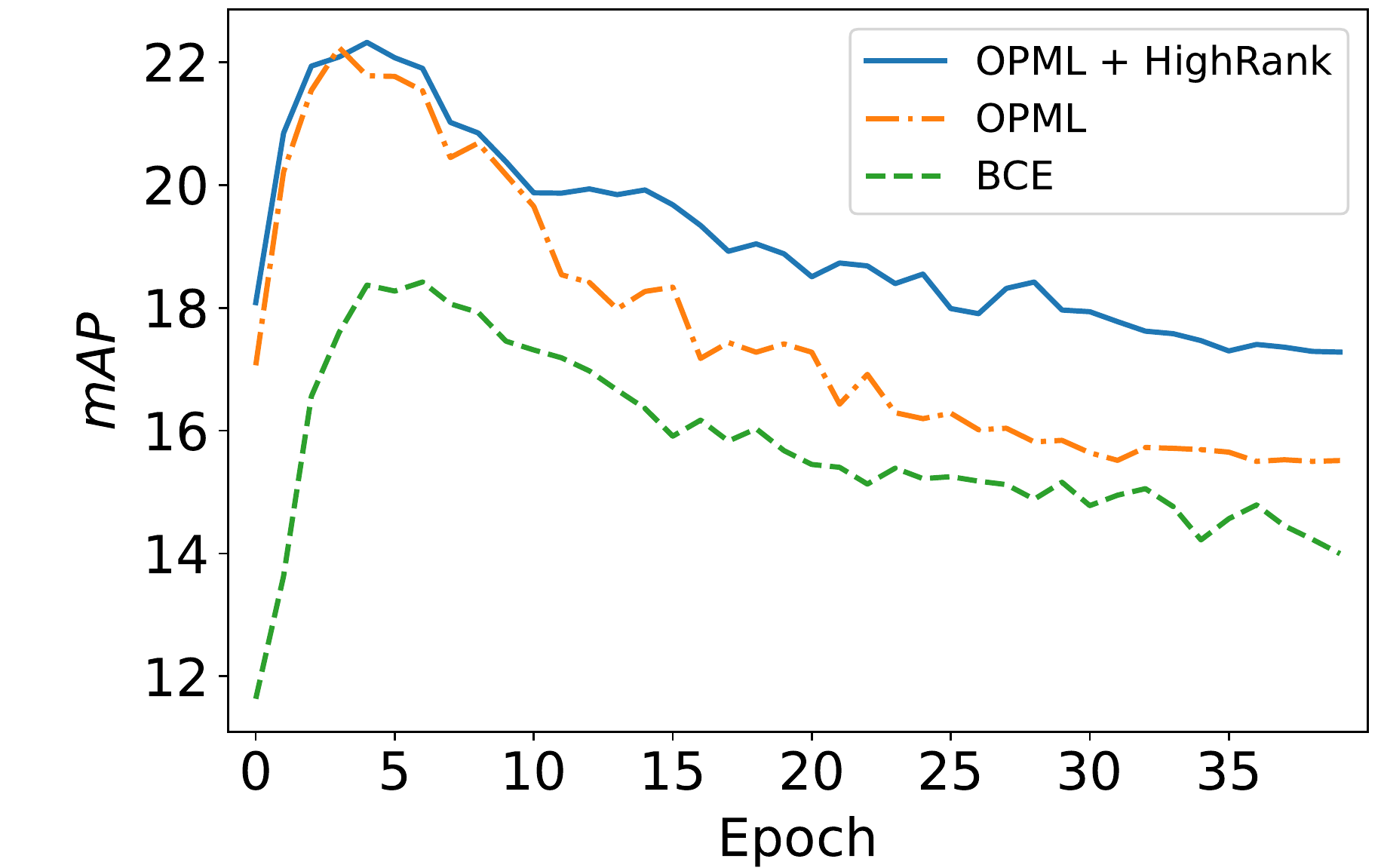}}				
	\caption{The effectiveness of high-rank regularization. The performance drop is not that dramatic with such a regularization.}
	\label{fig3}
\end{figure}

\noindent {\textbf {High-rank regularization.}} Compared with the first row in \Cref {tab4}, we can find that the mAP scores of the second row on all four datasets increase, which verifies the effectiveness of high-rank regularization. Besides, to validate the performance drop is not that dramatic with the high-rank regularization, we conduct experiments adopting BCE loss, OPML loss, and OPML loss with high-rank regularization, respectively. Results of running 40 epochs on VOC and CUB are shown in \Cref {fig3}. It can be seen that the mAP scores are more stable with the high-rank regularization. The reason may be that the high-rank property encourages to push different labels apart, which slows the process of turning the true positive labels to false negative labels. Thus, such a property reduces the impact of noisy label and achieves a stable training in SPMLL. We hope this discovering can bring new inspiration on general noisy label learning.  

\noindent {\textbf {Label smoothing and correction.}} From the third and the fourth rows of \Cref {tab4}, we can find that the performance is improved on all the four datasets with both the label smoothing and correction. Label smoothing achieves higher improvements than CUB on all the other three datasets, which can be attributed to that the ground-truth of CUB contains more positive labels, while label smoothing tries to transform the hard labels $\{0,1\}$ to continuous $[0,1]$, such an operation does not pay enough attention to the positive labels. Besides, label correction obtains noticeable improvement on all the four datasets, which validates that our training AP-based correction strategy is effective in SPMLL. To sum up, all the experiments illustrate that our OPML loss can be well cooperated with the label smoothing and correction.

\subsection{Hyper-parameters study}
\label{sec4.4}
\begin{figure} [h]
	\centering	
	\subfloat [mAP on VOC with different $\widetilde{\alpha}$ and $\widetilde{\beta}$.] {\label{fig4a} \includegraphics[width=0.473\linewidth]{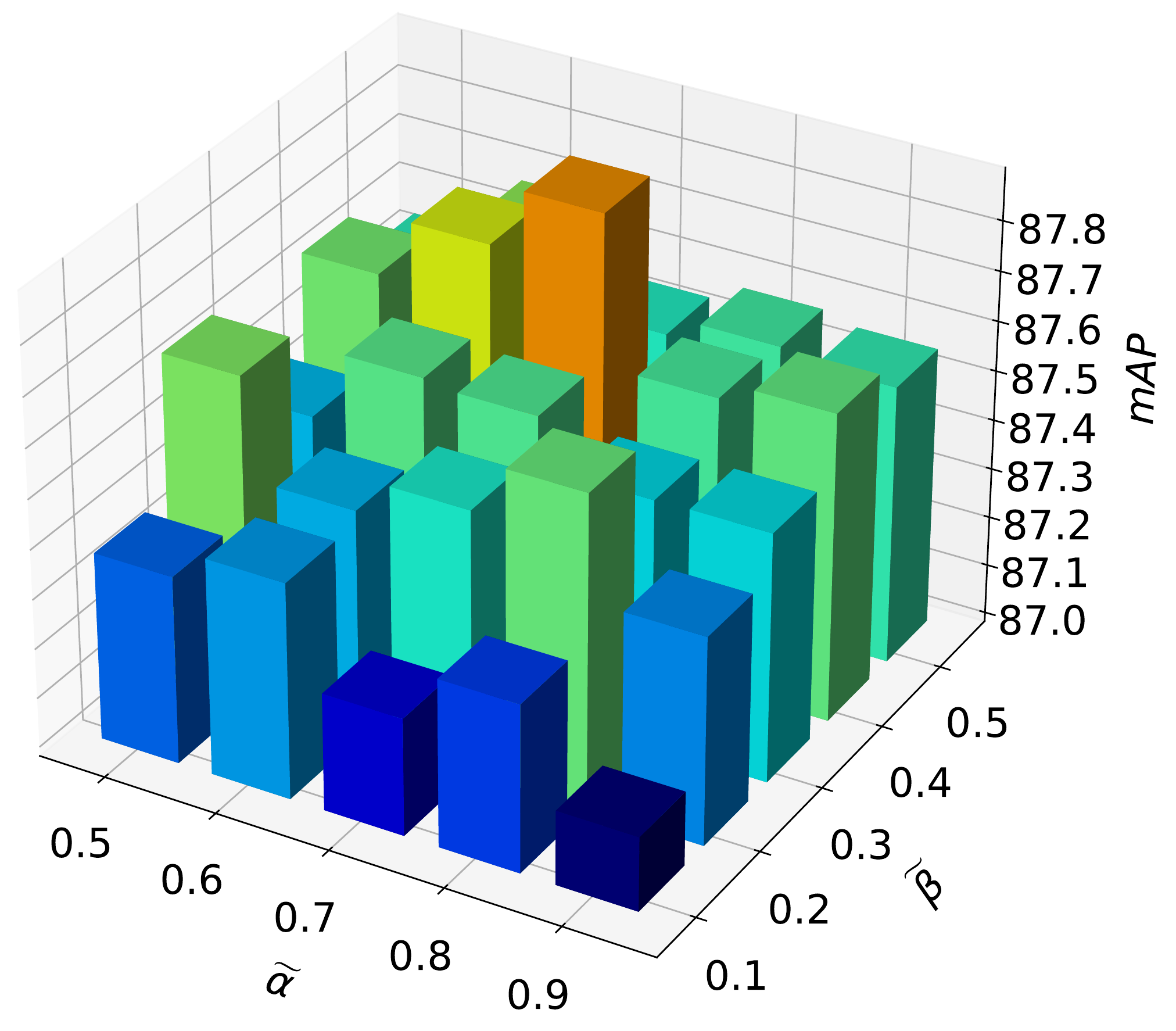}} \hspace {1mm}   			 
	\subfloat [mAP on CUB with different $\widetilde{\alpha}$ and $\widetilde{\beta}$.] {\label{fig4b} \includegraphics[width=0.473\linewidth]{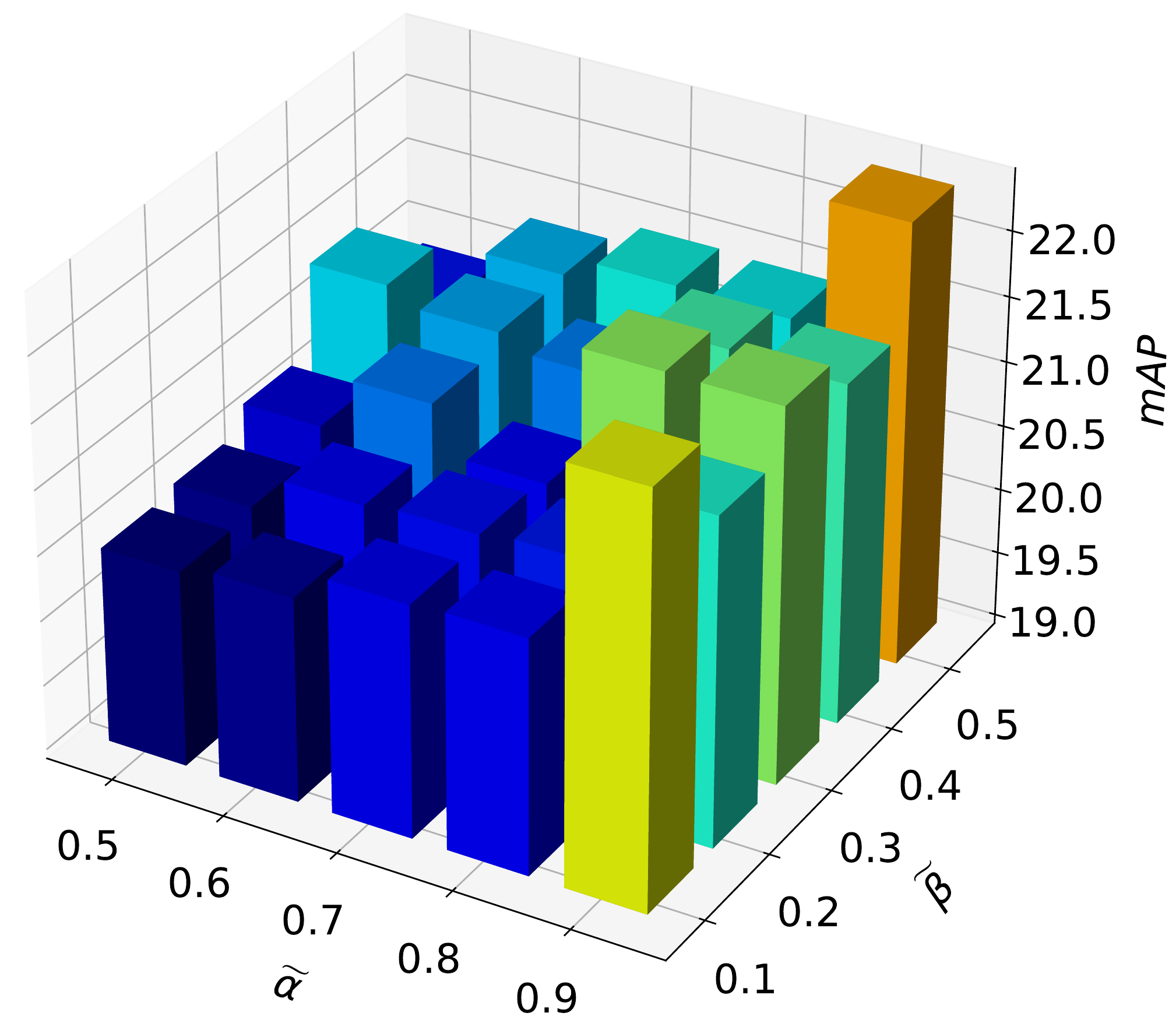}}\\
	\subfloat [mAP on COCO with different $\widetilde{\alpha}$ and $\widetilde{\beta}$.] {\label{fig4c} \includegraphics[width=0.473\linewidth]{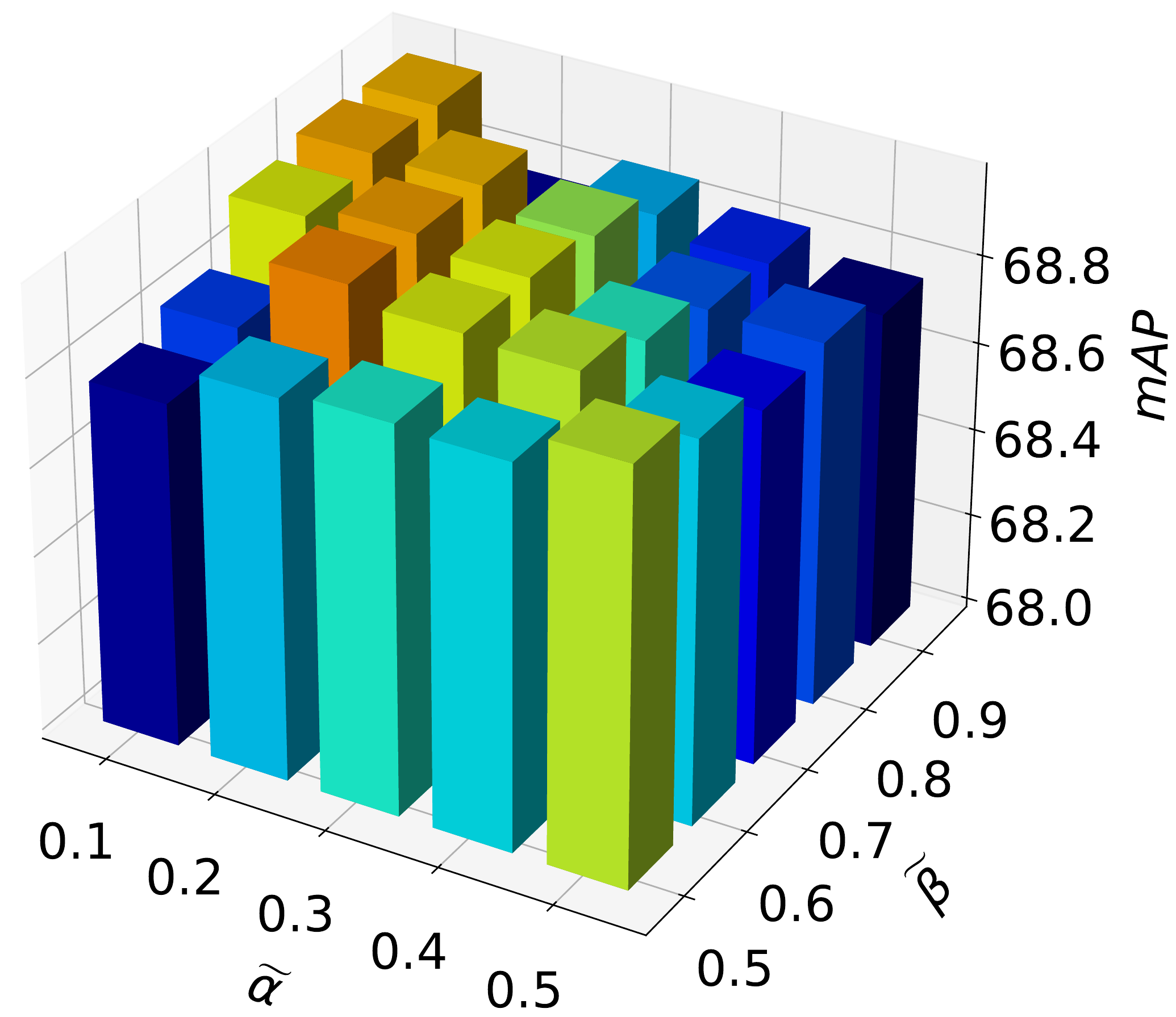}} \hspace {1mm} 			 
	\subfloat [mAP on NUS with different $\widetilde{\alpha}$ and $\widetilde{\beta}$.] {\label{fig4d} \includegraphics[width=0.473\linewidth]{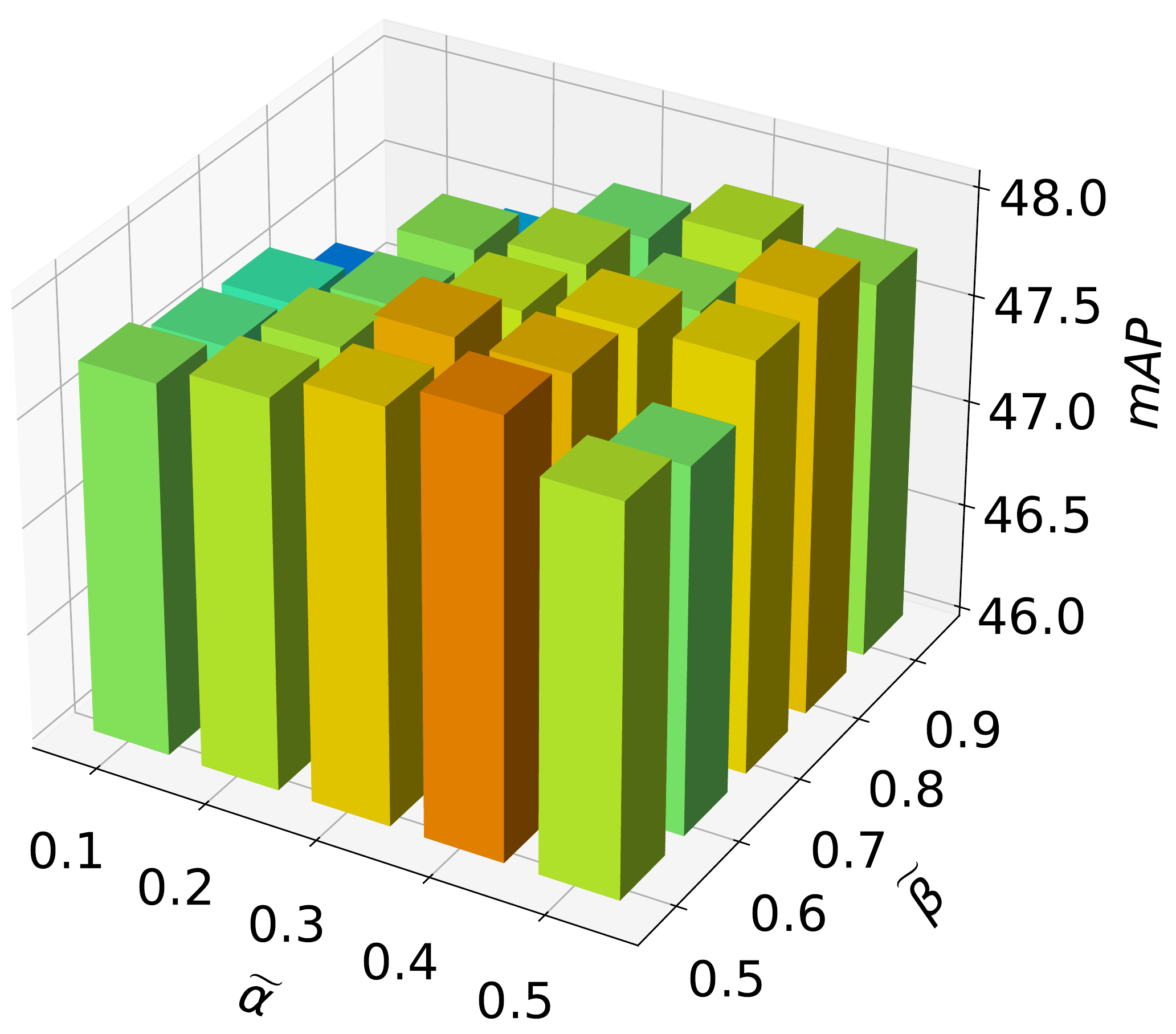}}	
	\caption{Hyper-parameters study of $\widetilde{\alpha}$ and $\widetilde{\beta}$. The best performance is marked with orange.}
	\label{fig4}
\end{figure}

In this subsection, we conduct experiments with different values of hyper-parameters to study their effects. Note that, the best hyper-parameter is selected by the best mAP on the validation set. In \Cref {fig4}, we show the results with different hyper-parameters of $\alpha$ and $\beta$ in \cref {eq:7}. Recalling that we transform $\alpha =  \widetilde{\alpha} / (1 - \widetilde{\alpha})$ and $\beta =  \widetilde{\beta} / (1 - \widetilde{\beta})$ for convenience of hyper-parameters selection, thus results are reported with different $\widetilde{\alpha}$ and $\widetilde{\beta}$. The best hyper-parameter is denoted by the orange pillar. $\widetilde{\alpha} = 0.7$ and $\widetilde{\beta}=0.4$ are the best parameters for VOC, and $\widetilde{\alpha} = 0.9$ and $\widetilde{\beta}=0.5$ are the best parameters for CUB. Note that, as a special case of our loss, the mAP values of ZLPR ($\widetilde{\alpha} = 0.5$ and $\widetilde{\beta}=0.5$) are lower than our OPML loss, which indicates that OPML is more flexible and achieves better performance. 

\begin{figure} [h]
	\centering	
	\subfloat [mAP with different $\lambda$ on VOC.] {\label{fig5a} \includegraphics[width=0.473\linewidth]{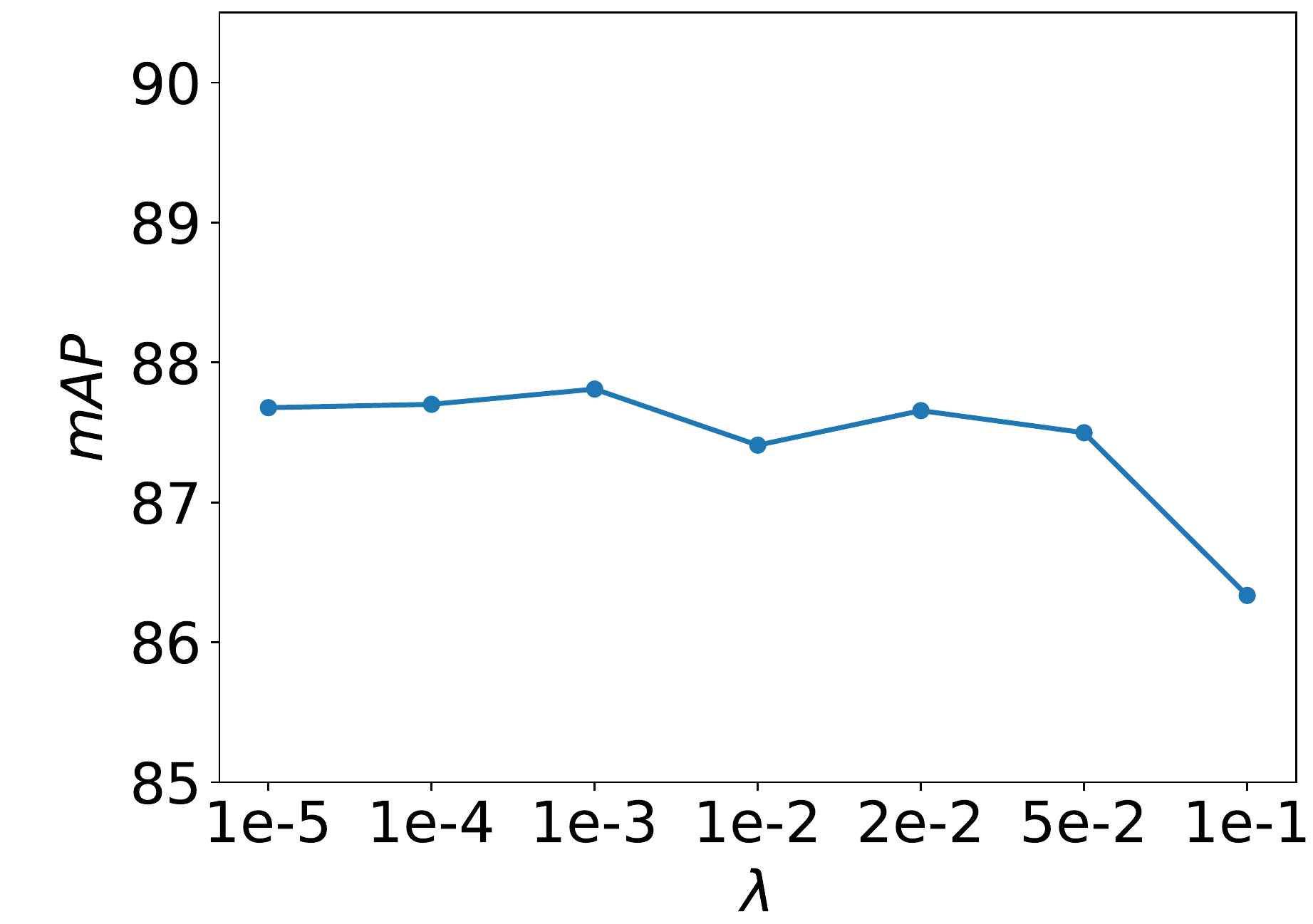}} \hspace {1mm} 				 
	\subfloat [mAP with different $\lambda$ on CUB.] {\label{fig5b} \includegraphics[width=0.473\linewidth]{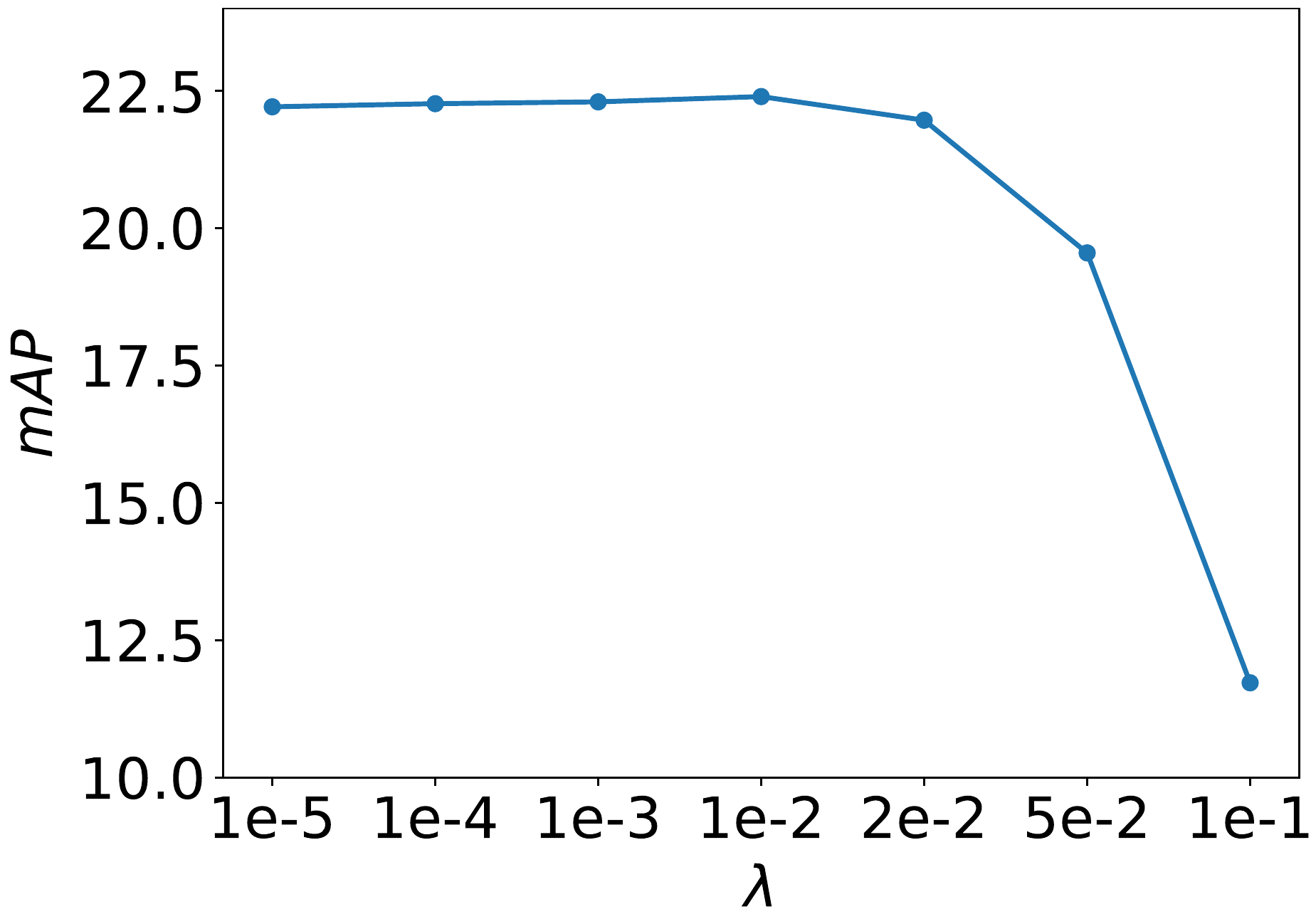}} \\
	\subfloat [mAP with different $\lambda$ on COCO.] {\label{fig5c} \includegraphics[width=0.473\linewidth]{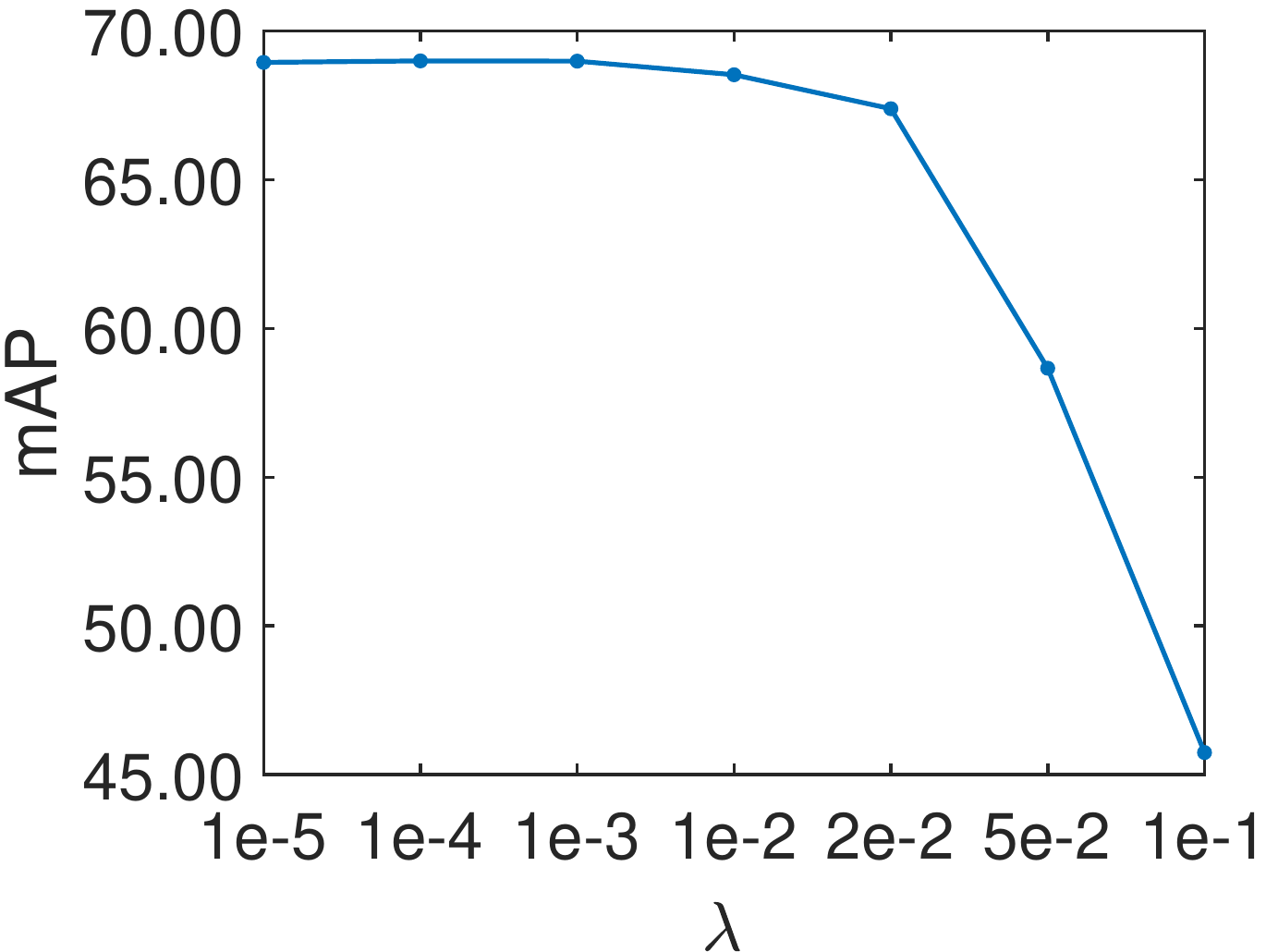}} \hspace {1mm} 				 
	\subfloat [mAP with different $\lambda$ on NUS.] {\label{fig5d} \includegraphics[width=0.473\linewidth]{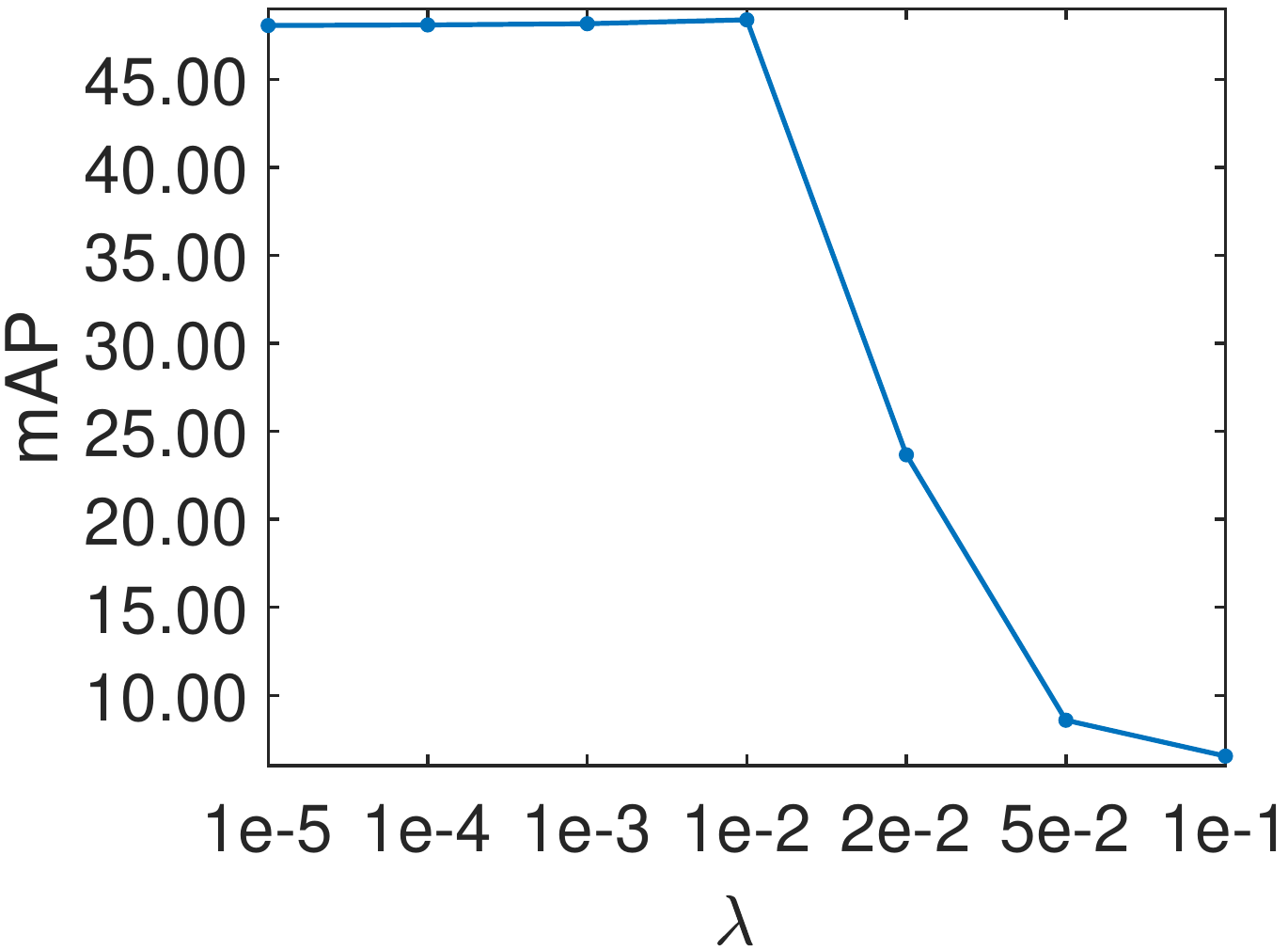}}
	\caption{The sensitivity study of $\lambda$. It is not sensitive to variations when less than $1e$-$2$ on both datasets.}
	\label{fig5}
\end{figure}

Besides, we also carry out experiments to study the sensitivity of the hyper-parameter $\lambda$ in \cref {eq:8}. To better observe the downtrend when $\lambda$ is larger than $1e$-$2$, we add the mAP values of $\lambda=2e$-$2$ and $\lambda=5e$-$2$. From \Cref {fig5}, we can see that the mAP value keeps stable when $\lambda$ is below $1e$-$2$, and the best $\lambda$ for VOC and CUB are $1e$-$3$ and $1e$-$2$, respectively. This figure demonstrates that $\lambda$ is not sensitive to variations when less than $1e$-$2$. 

\subsection{Grad-CAM visualization}
\label{sec4.5}
\begin{figure*}
	\centering
	\begin{minipage}{0.49\linewidth}
		\captionsetup[subfigure]{justification=centering}
		\subfloat[Ground truth:\\ horse, person]{\includegraphics[height=2.3cm, width=2.88cm]{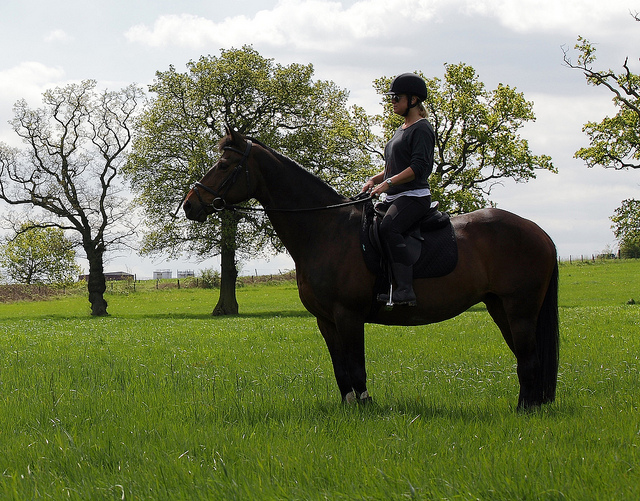}}
		\subfloat[horse]{\includegraphics[height=2.3cm, width=2.88cm]{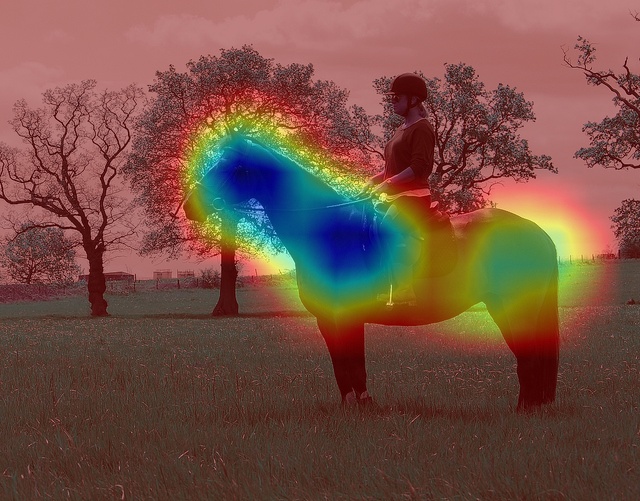}}
		\subfloat[person]{\includegraphics[height=2.3cm, width=2.88cm]{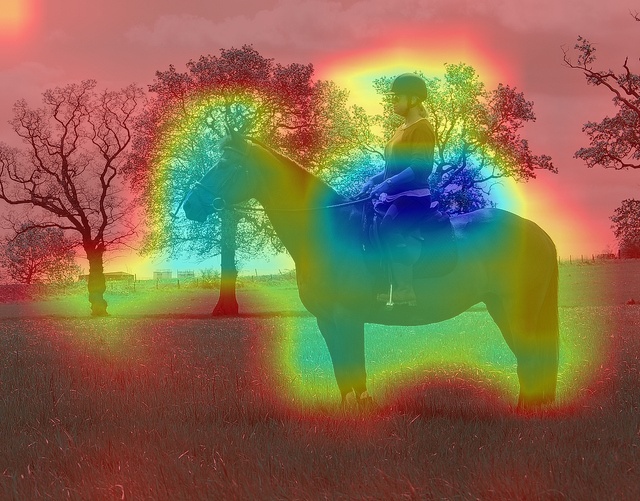}}
		\caption*{Figure 7-(1)}
	\end{minipage}
	\begin{minipage}{0.49\linewidth}
		\captionsetup[subfigure]{justification=centering}
		\setcounter{subfigure}{0}
		\subfloat[Ground truth: \\teddy bear, potted plant]{\includegraphics[height=2.3cm, width=2.88cm]{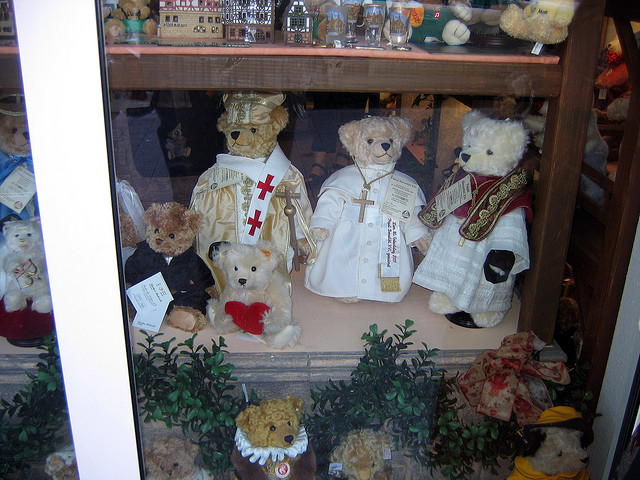}}
		\subfloat[teddy bear]{\includegraphics[height=2.3cm, width=2.88cm]{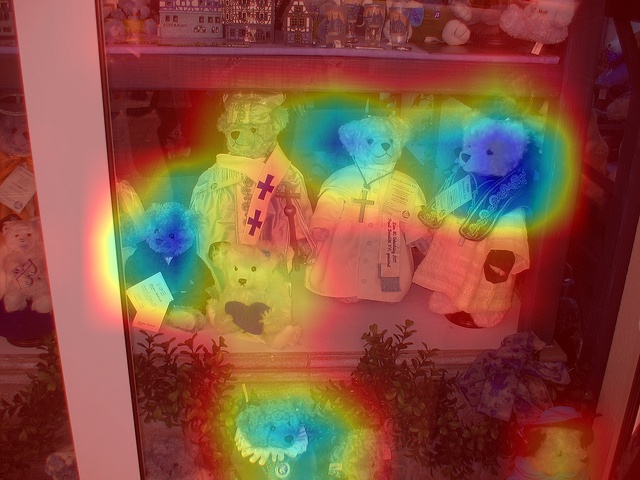}}
		\subfloat[potted plant]{\includegraphics[height=2.3cm, width=2.88cm]{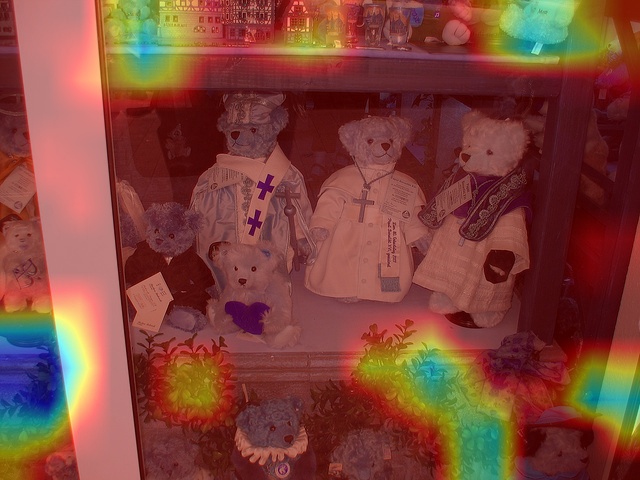}}
		\caption*{Figure 7-(2)}
	\end{minipage}
	
	\begin{minipage}{0.49\linewidth}
		\setcounter{subfigure}{0}
		\captionsetup[subfigure]{justification=centering}
		\subfloat[Ground truth: \\person, car]{\includegraphics[height=2.4cm, width=2.88cm]{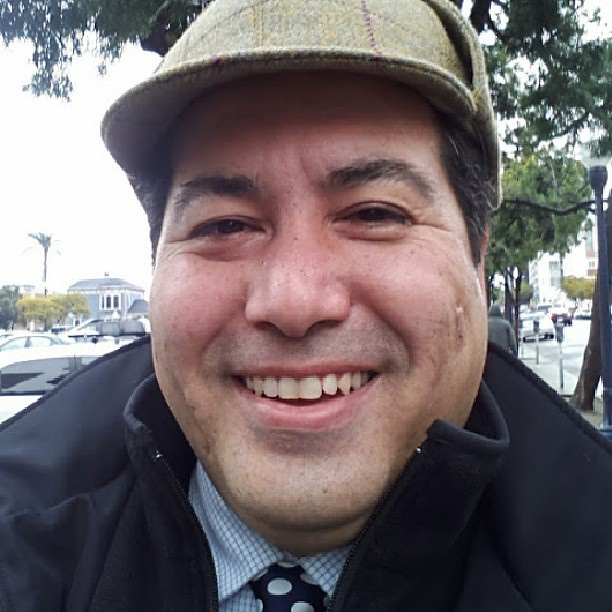}}
		\subfloat[person]{\includegraphics[height=2.4cm, width=2.88cm]{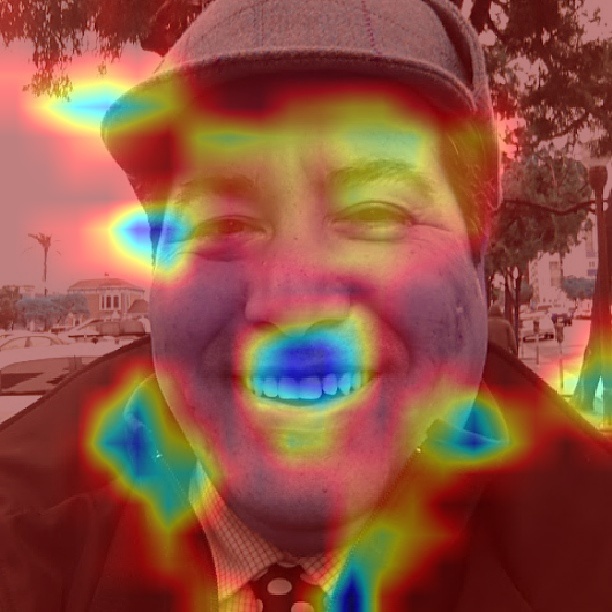}}
		\subfloat[car]{\includegraphics[height=2.4cm, width=2.88cm]{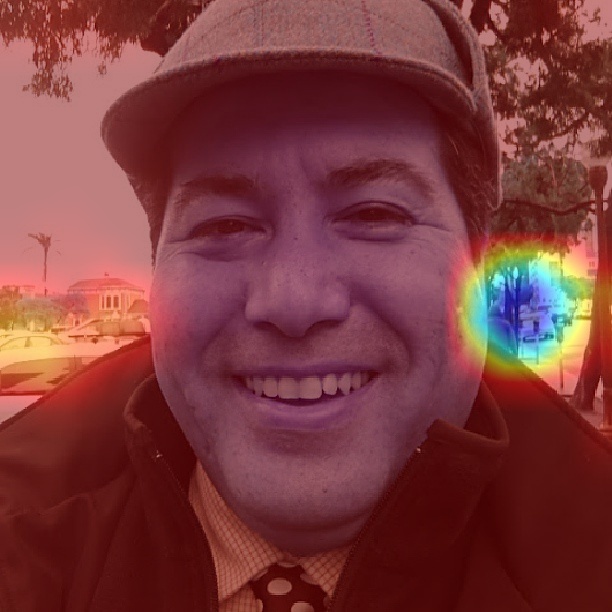}}
		\caption*{Figure 7-(3)}
	\end{minipage}
	\begin{minipage}{0.49\linewidth}
		\setcounter{subfigure}{0}
		\captionsetup[subfigure]{justification=centering}
		\subfloat[Ground truth: \\person, traffic light]{\includegraphics[height=2.4cm, width=2.87cm]{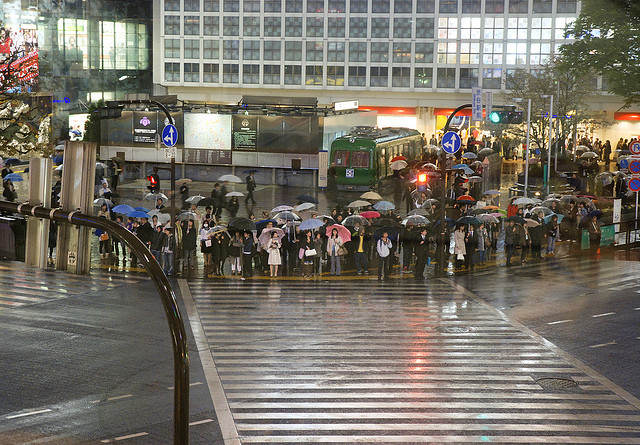}}
		\subfloat[person]{\includegraphics[height=2.4cm, width=2.87cm]{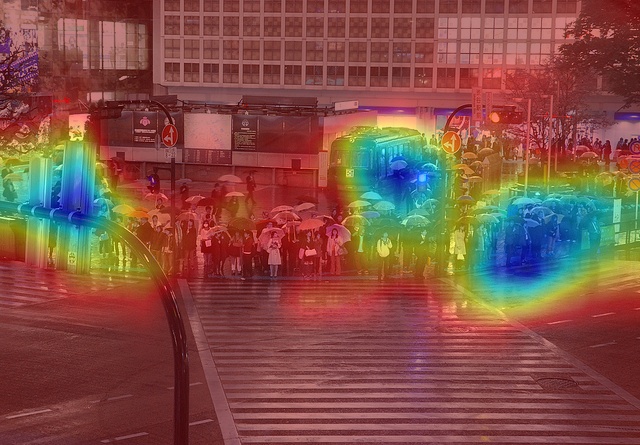}}
		\subfloat[traffic light]{\includegraphics[height=2.4cm, width=2.87cm]{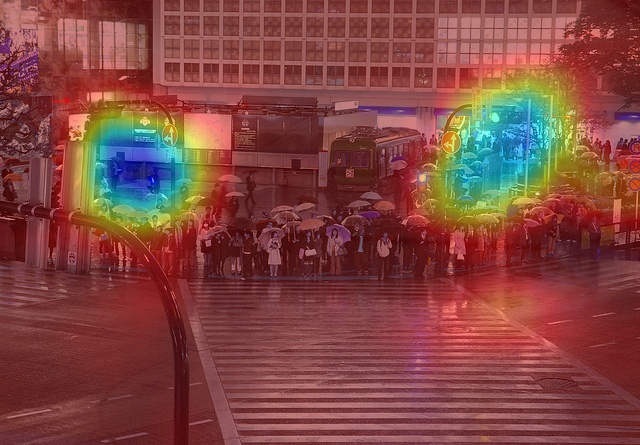}}
		\caption*{Figure 7-(4)}
	\end{minipage}
	
	\begin{minipage}{0.98\linewidth}
		\setcounter{subfigure}{0}
		\captionsetup[subfigure]{justification=centering}
		\subfloat[Ground truth: \\backpack, sofa, chair, book, potted plant]{\includegraphics[height=3cm, width=2.87cm]{3.jpg}}
		\subfloat[backpack]{\includegraphics[height=3cm, width=2.87cm]{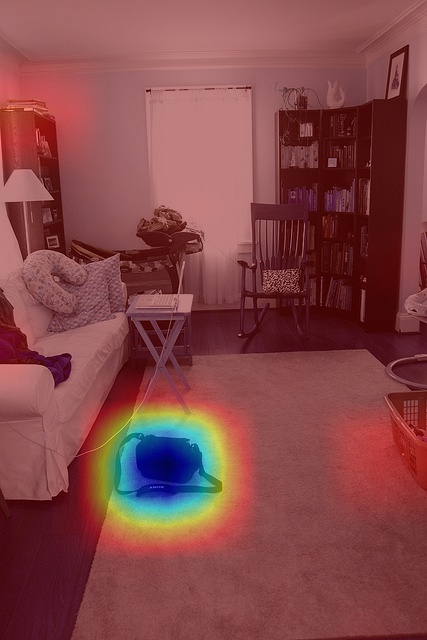}}
		\subfloat[sofa]{\includegraphics[height=3cm, width=2.87cm]{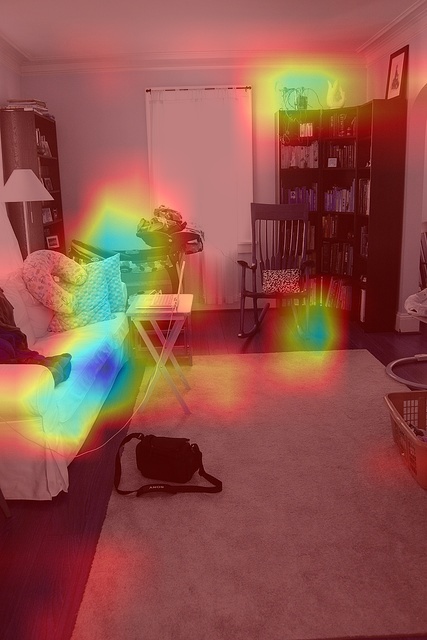}}
		\subfloat[chair]{\includegraphics[height=3cm, width=2.87cm]{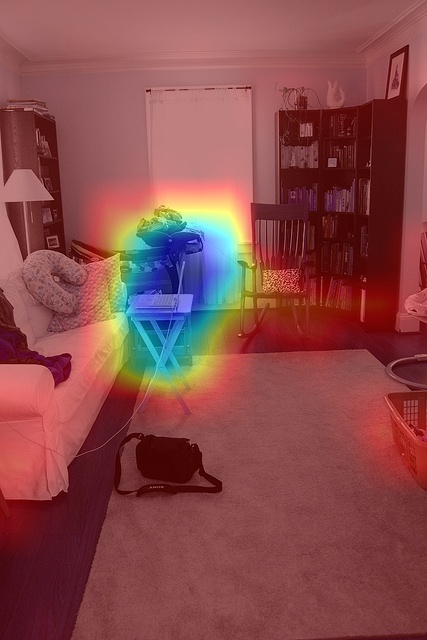}}
		\subfloat[book]{\includegraphics[height=3cm, width=2.87cm]{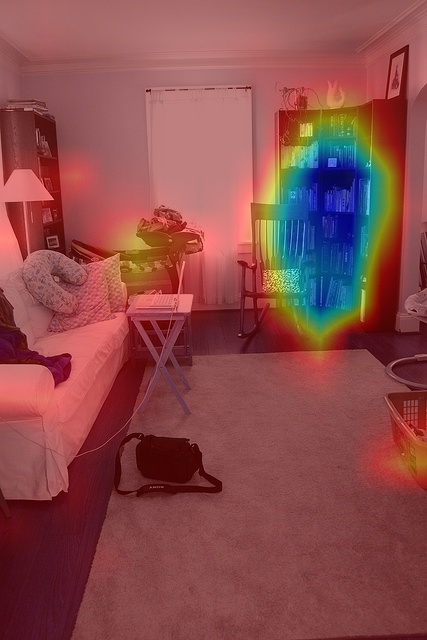}}
		\subfloat[potted plant]{\includegraphics[height=3cm, width=2.87cm]{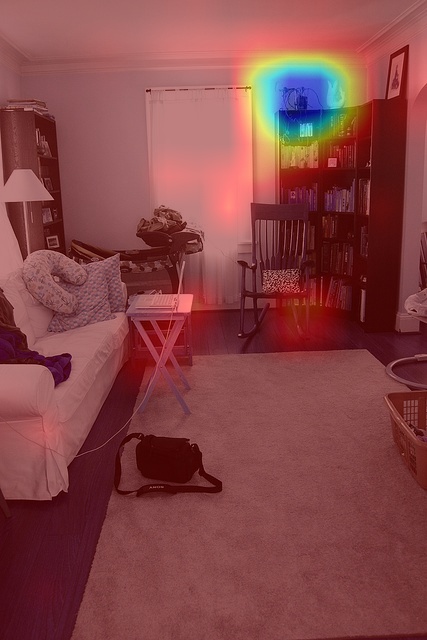}}
		\caption*{Figure 7-(5)}
	\end{minipage}
	
	\begin{minipage}{0.98\linewidth}
		\setcounter{subfigure}{0}
		\captionsetup[subfigure]{justification=centering}
		\subfloat[Ground truth: \\person, cup, cell phone]{\includegraphics[height=3cm, width=2.87cm]{12.jpg}}
		\subfloat[person]{\includegraphics[height=3cm, width=2.87cm]{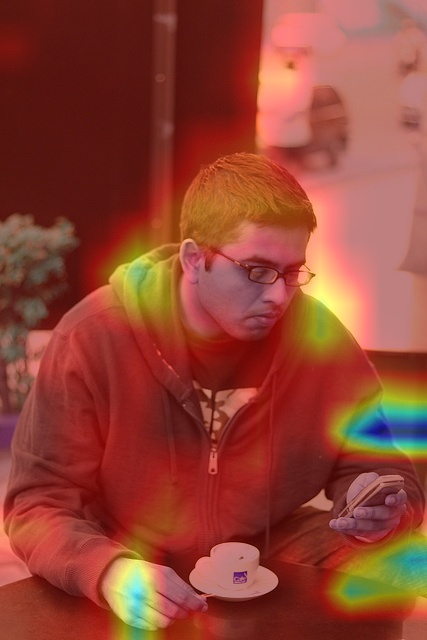}}
		\subfloat[cup]{\includegraphics[height=3cm, width=2.87cm]{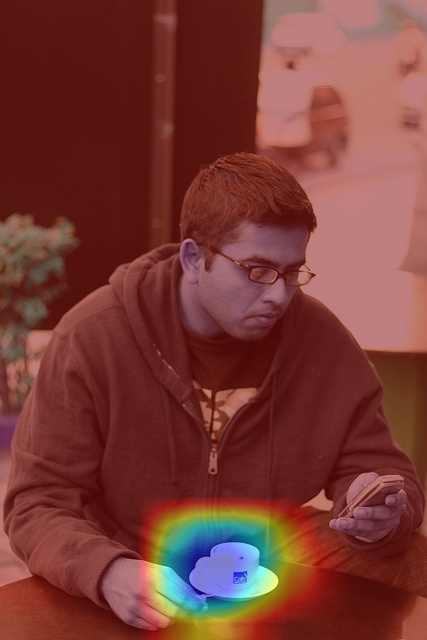}}
		\subfloat[cell phone]{\includegraphics[height=3cm, width=2.87cm]{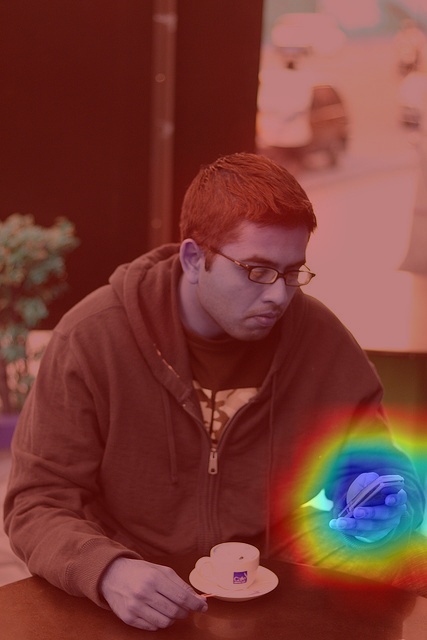}}
		\subfloat[\textit {\textbf {potted plant}}]{\includegraphics[height=3cm, width=2.87cm]{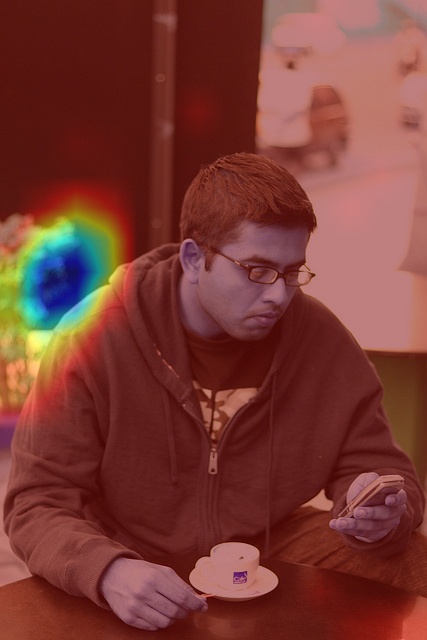}}
		\subfloat[\textit {\textbf {car}}]{\includegraphics[height=3cm, width=2.87cm]{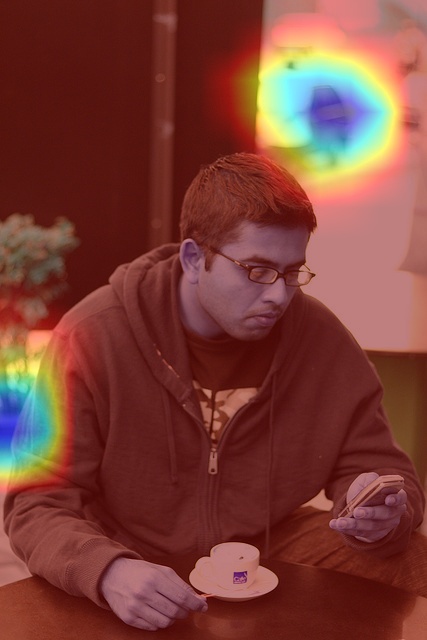}}
		\caption*{Figure 7-(6)}
	\end{minipage}
	
	\begin{minipage}{0.98\linewidth}
		\setcounter{subfigure}{0}
		\captionsetup[subfigure]{justification=centering}
		\subfloat[Ground truth: \\vase, chair, wine glass]{\includegraphics[height=2.5cm, width=2.16cm]{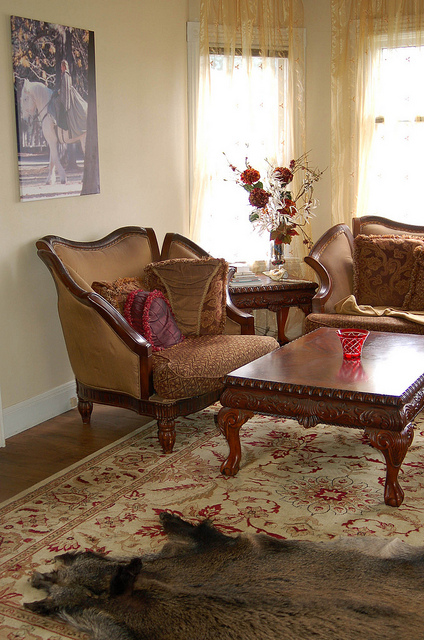}}
		\subfloat[vase]{\includegraphics[height=2.5cm, width=2.15cm]{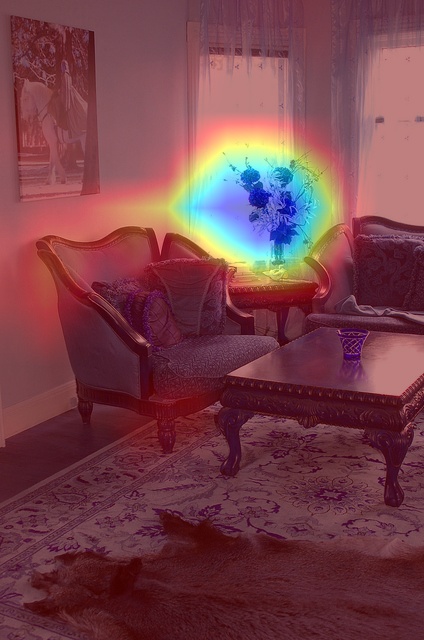}}
		\subfloat[chair]{\includegraphics[height=2.5cm, width=2.15cm]{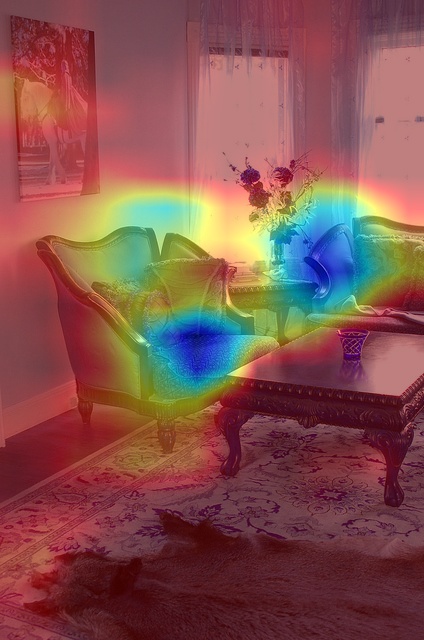}}
		\subfloat[wine glass]{\includegraphics[height=2.5cm, width=2.15cm]{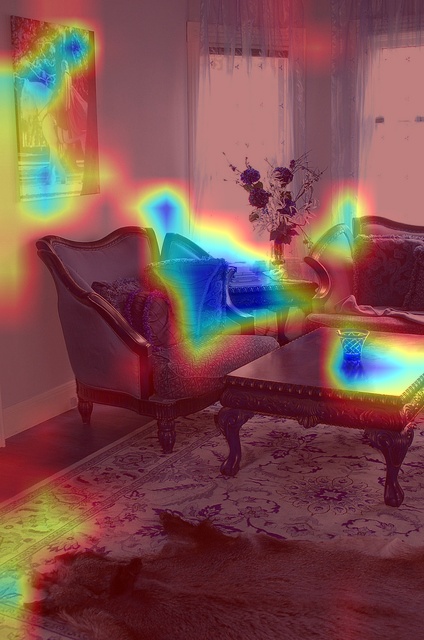}}
		\subfloat[\textit{\textbf {dining table}}]{\includegraphics[height=2.5cm, width=2.15cm]{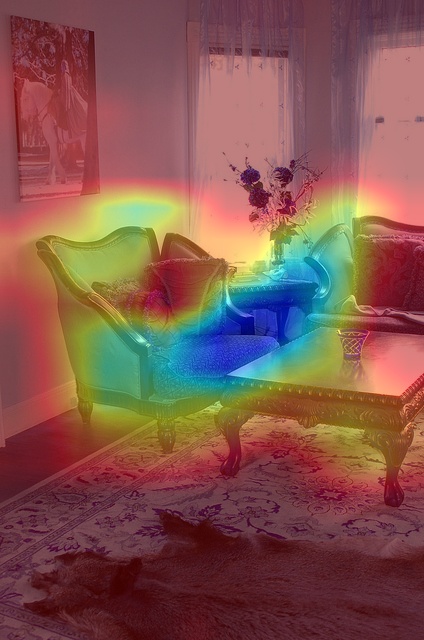}}
		\subfloat[\textit{\textbf {book}}]{\includegraphics[height=2.5cm, width=2.15cm]{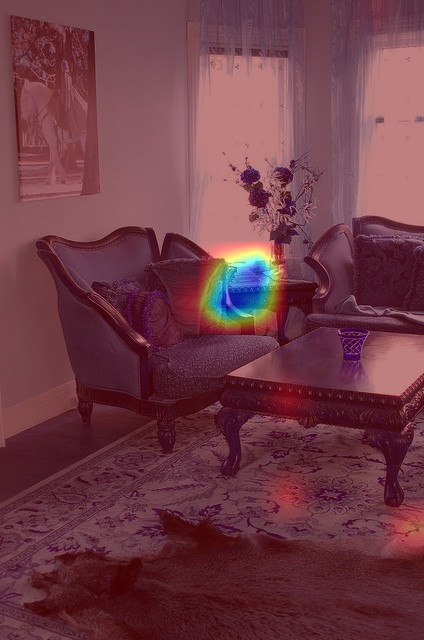}} 
		\subfloat[\textit{\textbf {person}}]{\includegraphics[height=2.5cm, width=2.15cm]{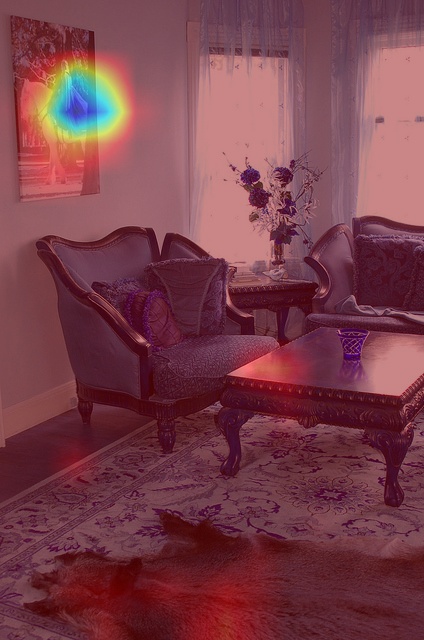}}
		\subfloat[\textit{\textbf {horse}}]{\includegraphics[height=2.5cm, width=2.15cm]{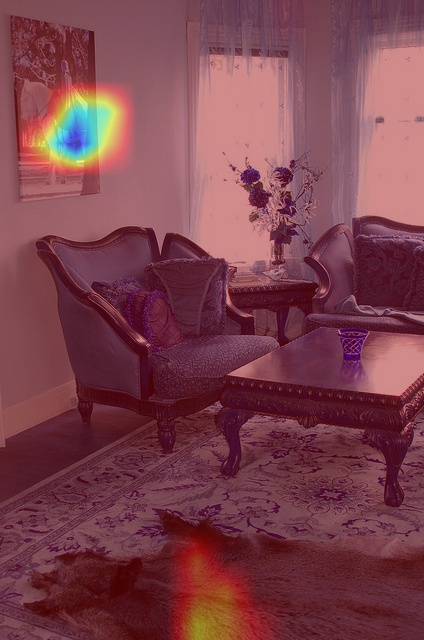}}
		\caption*{Figure 7-(7)}
	\end{minipage}
	\caption{Class Activation Mapping results of COCO test images. The blue color highlights the important regions in the image for predicting the corresponding concept. Lables in bold are that ignored by ground truth whereas discovered by our OPML loss.} 
	\label{fig3_new}
\end{figure*}

In this section, we utilize the Gradient-weighted Class Activation Mapping (Grad-CAM) \cite {selvaraju2017grad} to visualize the explanation of the proposed OPML loss, which produces a coarse localization map highlighting the important regions in the image for predicting the concept. In \Cref {fig3_new}, we show seven groups of pictures and their class activation maps, respectively. It can be seen that our OPML loss is capable of highlighting the locations of the concept to be recognized. It is worth mentioning that in Figure 7-(6) and 7-(7), the class activation maps discover missing concepts that are neglected in the ground truth but actually exist in the image. For example, in Figure 7-(6), the ground truth labels ignore the ``potted plant" and ``car", which indeed appear in the image. Especially for ``car", it can hardly be recognized even by human, and the same phenomenon can also be found in Figure 7-(7). These intriguing findings demonstrate that our OPML loss not only has an explanatory power that matches with humans, but also has the potential to help humans complete the missing labels, which can provide more accurate and complete labels for training.

\subsection{Experimental results of full labels}
In this subsection, we conduct experiments on MLL with full labels to validate that our OPML loss can still work well in such a setting, even though it is originated from the SPMLL setting. Note that, ASY \cite {ridnik2021asymmetric} is a state-of-the-art method in MLL with full labels, which focuses on addressing the imbalance between positive and negative labels. From \Cref {tab5}, we can find that our OPML loss performs the best on three datasets, which can be attributed to that pushing the positive label with the minimum score and negative label with the maximum score apart may increase the discrimination between them.
\begin{table}[!ht]
	\centering
	\small
	\setlength\tabcolsep{3pt}
	{
		\begin{tabular}{ccccc}
			\toprule[1.5pt]
			Methods & VOC & COCO & NUS & CUB \\ \midrule[0.8pt]
			BCE   & 89.42(0.27) & 76.78(0.13) & 52.08(0.20) & 30.90(0.64) \\ 
			FOCAL & 90.74(0.07) & \underline{77.35(0.13)} & 53.20(0.08) & 33.36(0.05) \\ 
			ASY   & 90.60(0.36) & \textbf{77.61(0.32)} & 52.84(0.17) & \underline{33.37(0.19)} \\ 
			ZLPR  & \underline{90.96(0.05)} & 76.35(0.10) & \underline{53.40(0.15)} & 33.27(0.12) \\ \midrule[0.8pt]
			OPML  & \textbf{91.36(0.06)} & 76.71(0.16) & \textbf{53.98(0.20)} & \textbf{33.72(0.08)} \\ \bottomrule[1.5pt]
		\end{tabular}
	}
	\caption{Quantitative results of the commonly used loss functions in MLL and our OPML loss on four benchmarks with full labels. The mean and standard deviation of mAP are reported. The best result is in bold and the second best result is underlined.}
	\label{tab5}
\end{table}

\subsection{Details of experimental implementation}
To better reproduce the experimental results reported in this work, we elaborate the details of experimental implementation in this section. The Stochastic Gradient Descent (SGD) optimizer is used for training. The parameters of OPML-SP (the abbreviation of OPML loss for Single Positive setting) on dataset VOC are listed as follows: batch size is $8$, learning rate is $1e-4$, running epochs are $15$, $\widetilde{\alpha} = 0.6$, $\widetilde{\beta} = 0.4$, smoothing power parameter $\epsilon = 0.7$, label correction parameter $Label_{num}=0.8$. The parameters used on COCO are that batch size is $16$, learning rate is $2e-4$, running epochs are $10$, $\widetilde{\alpha} = 0.4$, $\widetilde{\beta} = 0.5$, smoothing power parameter $\epsilon = 1$, and label correction parameter $Label_{num}=1.2$. Besides, the parameters used on NUS are that batch size is $128$, learning rate is $1e-3$, running epochs are $10$, $\widetilde{\alpha} = 0.6$, $\widetilde{\beta} = 0.4$, smoothing power parameter $\epsilon = 1$, label correction parameter $Label_{num}=1.1$. Finally, the parameters used on CUB are as below: batch size is $8$, learning rate is $4e-4$, running epochs are $10$, $\widetilde{\alpha} = 0.6$, $\widetilde{\beta} = 0.2$, smoothing power parameter $\epsilon = 1$, and the label correction parameter $Label_{num}=22$. The hyper-parameter $\lambda$ is fixed at $1e-3$ on all four datasets. Note that, the parameters reported here are not fine-tuned, and performance may be further improved with fine-tuned parameters.

\section {Conclusion}
In this paper, we present a novel unified loss named OPML for both SPMLL and MLL with full labels by pushing one pair of labels apart each time to prevent the domination of negative labels. Experiments on four benchmarks verify that the OPML loss not only performs more robustly in SPMLL for alleviating the impact of noisy labels but also works well in MLL with full labels for separating the positive and negative labels. Besides, we empirically find that the high-rank property of the label matrix can slow down the dramatic performance drop, which may shed new light on general noisy label learning. Note that, the imbalance between the positive and negative labels becomes more severe in SPMLL, thus, how to deal with this issue may be a future research direction for further closing the gap between SPMLL and MLL with full labels.

\normalem
\bibliographystyle{IEEEtran}
\bibliography{IEEEabrv,egbib}

\begin{IEEEbiography}[{\includegraphics[width=1in,height=1.25in,clip,keepaspectratio]{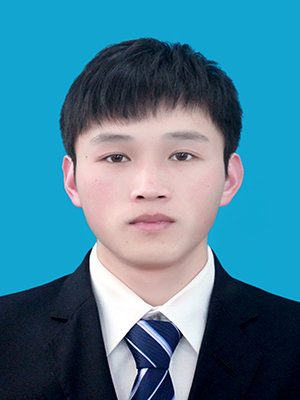}}]{Xiang Li}
	received his B.S. degree in mathematics from College of Science, Nanjing University of Aeronautics and Astronautics in 2012. And he completed his M.S. degree in applied mathematics from College of Science, Nanjing University of Aeronautics and Astronautics in 2018. 
	
	Now he is a fourth-year Ph.D. candidate in College of Computer Science and Technology, Nanjing University of Aeronautics and Astronautics. His research interests include multi-view learning and multi-label learning.
\end{IEEEbiography}

\begin{IEEEbiography}[{\includegraphics[width=1in,height=1.25in,clip,keepaspectratio]{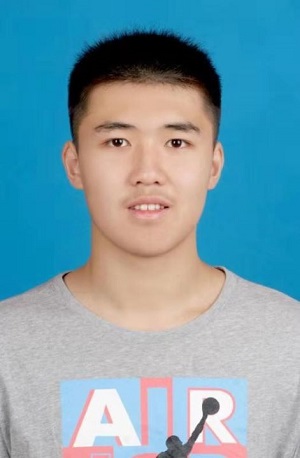}}]{Xinrui Wang}
	received his B.S. degree in Computer Science from College of Computer Science, Northwestern Polytechnical University, 2021. 
	
	Now he is a first-year Ph.D. candidate in College of Computer Science and Technology, Nanjing University of Aeronautics and Astronautics. His research interests include semi-supervised learning and multi-label learning.
\end{IEEEbiography}

\begin{IEEEbiography}[{\includegraphics[width=1in,height=1.25in,clip,keepaspectratio]{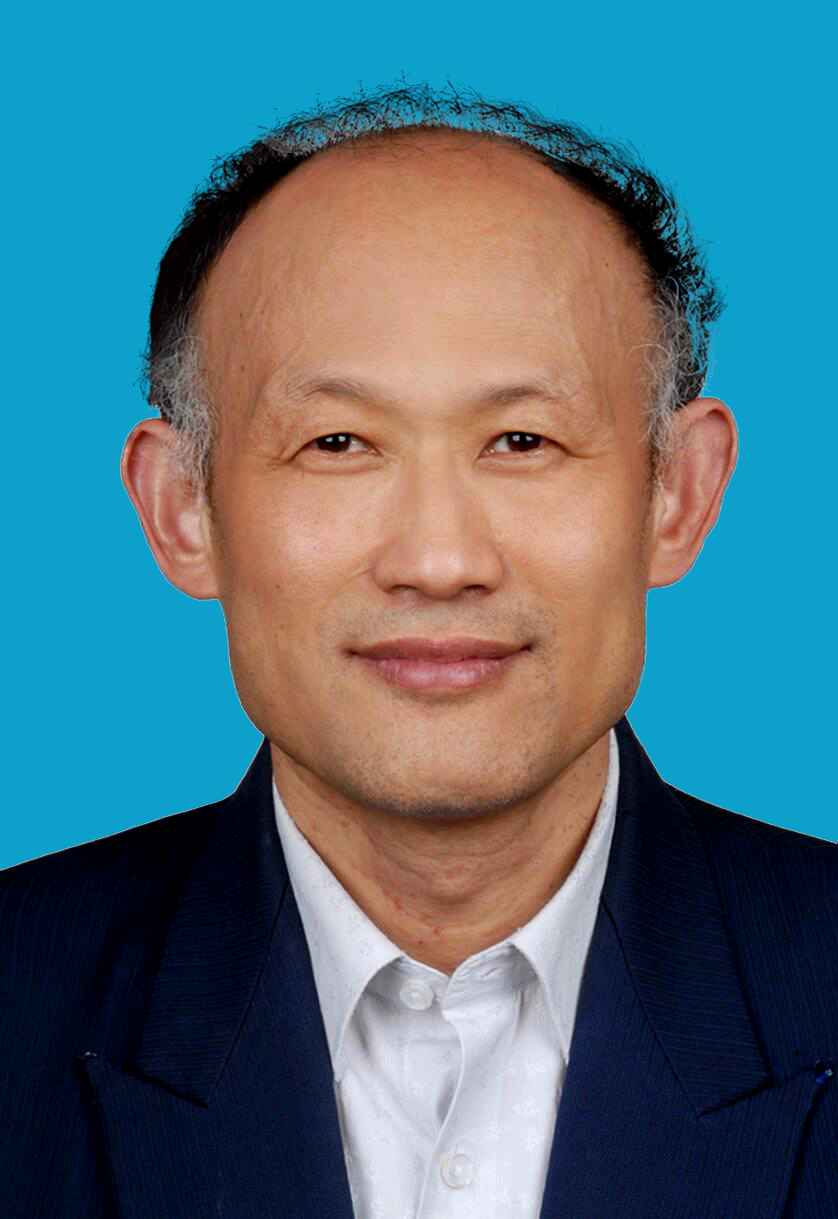}}]{Songcan Chen}
	received his B.S. degree in mathematics from Hangzhou University (now merged into Zhejiang University) in 1983. In 1985, he completed his M.S. degree in computer
	applications at Shanghai Jiaotong University and then worked at Nanjing University of Aeronautics and Astronautics (NUAA) in January 1986, where he received a Ph.D. degree in communication and information systems in 1997.
	
	Since 1998, as a full-time professor, he has been with the College of Computer Science and Technology at NUAA. His research interests include pattern recognition, machine learning and neural computing. He has published over 100 top-tier journals, such as IEEE Transactions on Pattern Analysis and Machine Intelligence, IEEE Transactions on Knowledge and Data Engineering, IEEE Transactions on Image Processing, IEEE Transactions on Neural Networks and Learning Systems, IEEE Transactions on Fuzzy Systems, IEEE Transactions on Information Forensics \& Security, IEEE Transactions on Systems, Man \& Cybernetics-Part B, IEEE Transactions on Wireless Communication and conference papers, such as International Conference on Machine Learning, IEEE Conference on Computer Vision and Pattern Recognition, International Joint Conference on Artificial Intelligence, AAAI Conference on Artificial Intelligence, IEEE International Conference on Data Mining and so on. He is also an IAPR Fellow.
\end{IEEEbiography}

\end{document}